\documentclass{article}

\PassOptionsToPackage{numbers, compress}{natbib} 

 \usepackage[preprint]{neurips_2026}

\usepackage[utf8]{inputenc} 
\usepackage[T1]{fontenc}    
\usepackage{url}            
\usepackage{booktabs}       
\usepackage{amsfonts}       
\usepackage{nicefrac}       
\usepackage{microtype}      
\usepackage[dvipsnames,table]{xcolor}
\usepackage[colorlinks,
            linkcolor=red,
            anchorcolor=blue,
            citecolor=green
            ]{hyperref}
\usepackage{caption}
\usepackage{subcaption}
\usepackage{natbib}
\usepackage{graphicx}
\usepackage{enumitem}
\usepackage{makecell}
\usepackage{amsmath}
\usepackage{amsthm}
\usepackage{thmtools}
\usepackage{thm-restate}
\usepackage{algorithm}
\usepackage{algorithmic}
\usepackage{bbm}
\usepackage{bm}
\usepackage{bbding} 
\usepackage{multirow}
\usepackage[most,breakable]{tcolorbox}
\usepackage{verbatim}
\usepackage{fancyvrb}
\usepackage{booktabs}

\usepackage{eso-pic}
\newcommand{\longcatlogoposfirst}{\AtPageUpperLeft{\hspace{37.7mm}\raisebox{-26.3mm}{\includegraphics[height=9mm]{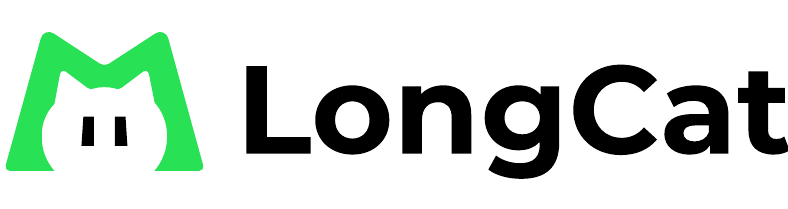}}}}

\AddToShipoutPictureFG{%
  \ifnum\value{page}=1\relax
    \longcatlogoposfirst
  \fi
}

\usepackage{tocloft}
\hypersetup{
    colorlinks=true,
    linkcolor=red,
    filecolor=magenta,
    urlcolor=cyan,
    citecolor=cyan,
}

\definecolor{-}{rgb}{0.25,0.41,0.88}
\definecolor{+}{rgb}{0.70,0.13,0.13}

\title{VitaBench 2.0: Evaluating Personalized and Proactive Agents in Long-Term User Interactions}

\author{
  \textbf{Yuxin Chen$^{1,2}$},
  \textbf{Yi Zhang$^{2,3}$},
  \textbf{Zhengzhou Cai$^{2,4}$},
  \textbf{Yaorui Shi$^{2,3}$},
  \textbf{Zhiyuan Yao$^{2,5}$},
  \\
  \textbf{Chenhang Cui$^{1,2}$},
  \textbf{Jingnan Zheng$^{1,2}$},
  \textbf{Yaqi Huo$^{2}$},
  \textbf{Xi Su$^{2}$},
  \\
  \textbf{Qi Gu$^{2,\dagger}$},
  \textbf{Xunliang Cai$^{2}$},
  \textbf{Xiang Wang$^{3}$},
  \textbf{An Zhang$^{3,\dagger}$},
  \textbf{Tat-Seng Chua$^{1}$}
  \\
  \vspace{-2mm} \\
  $^{1}$National University of Singapore,
  $^{2}$Meituan, \\
  $^{3}$University of Science and Technology of China, \\
  $^{4}$Beijing University of Posts and Telecommunications,
  $^{5}$Zhejiang University \\
  $^{\dagger}$Corresponding authors: \texttt{guqi03@meituan.com}, \texttt{an\_zhang@ustc.edu.cn}
}

\begin{document}

\maketitle

\begin{abstract}
\label{abstract}

Large language models (LLMs) have evolved into interactive agents that collaborate with users in real-world tasks.
Effective collaboration in such settings increasingly depends on understanding the user beyond what is explicitly stated, as user intent is often reflected in fragmented daily interactions and requires both personalized modeling and proactive interaction.
However, existing agent benchmarks primarily evaluate reasoning and tool use, largely overlooking the challenges of inferring and leveraging user preferences in realistic scenarios.
To address this gap, we introduce VitaBench 2.0, a benchmark for evaluating personalized and proactive agent behavior in long-term user interactions.
In VitaBench 2.0, tasks are organized as temporally ordered sequences for individual users, where preferences are embedded in fragmented and heterogeneous interactions.
Successful completion of tasks requires the agent to continuously extract, utilize, and update user preferences from these interactions. 
We further evaluate proactiveness through tasks that require agents to recognize missing information and actively acquire it from users or environments before making decisions.
To support systematic analysis, we provide an extensible memory interface that enables controlled comparison across different memory architectures.
We benchmark a diverse set of frontier proprietary and open-source LLMs. 
Results show that real-world personalization remains highly challenging even for state-of-the-art models, revealing a substantial gap between current capabilities and practical requirements.
Extensive analysis further reveals the failure modes and capability bottlenecks of current agents in real-world personalized decision-making, providing insights for future model improvements.
Code is available at \url{https://github.com/meituan-longcat/vitabench-2.0}.
\end{abstract}
\addtocontents{toc}{\protect\setcounter{tocdepth}{-1}}
\section{Introduction}
\label{introduction}

Recent advances in large language models (LLMs) have improved their capabilities in reasoning and tool use~\cite{deepseekv31,guo2025deepseekr1,gpt5,anthropic2025claude45}, enabling them to evolve from passive text generators into interactive agents operating in real-world environments~\cite{longcatflash2601, liu2025deepseekv32,qwen3max}. 
As these agents move from single-turn interactions to sustained collaboration with users, effective assistance increasingly depends on understanding user intent beyond what is explicitly stated~\cite{liu2025survey_personalized_llm}. 
In real-life scenarios, such intent is often reflected implicitly through fragmented interactions~\cite{jiang2025personamem_v2, jiang2025personamem,chen2026knowu}, making personalization central to user–agent collaboration.

However, this growing need for personalization in human–agent collaboration remains insufficiently captured by existing agent benchmarks.
Existing benchmarks primarily focus on evaluating multi-step reasoning and tool orchestration, where tasks are well-specified and the context required for successful completion is clearly stated within the context~\cite{jimenez2024swebench,liu2023agentbench, zhou2023webarena,yao2024tau_bench, barres2025tau2_bench, he2025vitabench}.
As a result, they mainly evaluate agents’ ability to follow explicit instructions and execute correct action sequences.
In contrast, emerging real-world agent systems increasingly operate in settings where user intent is under-specified and must be inferred from prior interactions~\cite{zhang2024personalization_survey}.
In such scenarios, effective assistance requires agents to maintain a consistent representation of user preferences, adapt to their evolution over time, and proactively acquire missing information when necessary.
This shift introduces a fundamentally different source of complexity, moving beyond reasoning over explicit instructions to decision-making grounded in implicit and evolving user preferences.
This gap highlights the need for agent benchmark that explicitly evaluates personalization and proactiveness in realistic user-agent interaction settings.

Toward this end, we introduce VitaBench 2.0, an agent benchmark for evaluating personalized and proactive behavior in real-world long-term user interactions.
Beyond tool use and reasoning ability, VitaBench 2.0 also evaluates personalization along three dimensions: (1) \textit{preference extraction}, where agents infer implicit preferences from fragmented interactions; (2) \textit{preference utilization}, where agents leverage these preferences for user-specific decision-making; and (3) \textit{preference updating}, where agents capture preference drift and revise their understanding as user behavior evolves.
Building on this formulation, we further evaluate \textit{proactiveness}, which arises when user preference is conditional and requires agents to actively acquire missing information before making decisions.

Following the general setup of existing agent benchmarks~\cite{yao2024tau_bench,barres2025tau2_bench,he2025vitabench}, VitaBench 2.0 is constructed as an interactive agent benchmark, where agents interact with environments to fulfill user needs.
Tasks in VitaBench 2.0 are organized as temporally ordered sequences for individual users, where each task sequence spans multiple domains, and each task is paired with a dedicated set of tools and an executable environment to support realistic interaction.
To evaluate personalization, we curate a series of fine-grained preferences for each user and embed them into fragmented interactions, including both dialogues and behaviors.
As agents continuously interact with users over time, user preferences may evolve, which is reflected in newly observed interactions, requiring agents to maintain and update a consistent representation of preferences within task sequences.
To capture long-term user dynamics in realistic interaction settings, we allow agents to maintain a memory module for each user.
Building on this, VitaBench 2.0 provides an extensible memory interface that supports flexible implementations and enables controlled comparison across representative memory mechanisms~\cite{mem0_2024,amem,memagent}.

We conduct extensive evaluations on a wide range of frontier proprietary and open-source language models. 
Our results show that real-world personalization tasks remain highly challenging for current agents, revealing a substantial gap between existing capabilities and practical requirements. 
We further analyze the role of memory and find that, while memory mechanisms are essential for long-term user modeling, existing approaches often fail to consistently translate stored information into improved performance, and different memory designs lead to markedly different outcomes.
Through systematic analysis, we identify key failure patterns and primary bottlenecks of current agents, providing insights into why current models struggle with personalization.
VitaBench 2.0 highlights a gap between current LLM agents and realistic personalized assistants and provides a testbed for future research on memory, personalization, and proactive agent behavior.

\section{Related Work} 
\label{sec:related-work}

\paragraph{Personalized LLM.}
As large language models are increasingly deployed in user-facing applications, personalization has become a critical capability for aligning model outputs with individual user needs and preferences~\cite{liu2025survey_personalized_llm,zhang2024personalization_survey,tseng2024two_tales_persona}. Achieving personalization requires models to capture user-specific characteristics and incorporate them into the generation process. Existing methods can be broadly understood from three alignment perspectives: input-level alignment, model-level alignment, and objective-level alignment.
Input-level alignment enriches prompts with user-specific context. Retrieval-augmented methods obtain such context from interaction histories or external knowledge stores~\cite{salemi2024optimization_ropg,mysore2023pearl,richardson2023pag}, while profile-based approaches explicitly summarize and inject user preferences into the prompt~\cite{li2023teach_llm_personalize,wu2024understanding_user_profile}. Model-level alignment adapts the model itself to generate outputs conditioned on user preferences, through parameter adaptation for white-box models~\cite{tan2024oppu,zhang2024plora} or model factorization frameworks for black-box models~\cite{zhuang2024hydra}. Objective-level alignment incorporates personalization into training objectives, including personalized reward modeling~\cite{jang2023personalized_soups}, multi-objective preference optimization~\cite{zhou2023modpo}, and causal preference modeling~\cite{zhao2025nextquill}.
As user interactions become increasingly long-term and informative, memory-augmented personalization has gained growing attention, supported by advances in memory systems and the increasing capability of LLMs to utilize them. This line of work augments LLMs with external memory mechanisms that support the storage, retrieval, and updating of user-relevant information over time~\cite{packer2023memgpt,mem0_2024,xu2025amem}.

\paragraph{Benchmarks for LLM Personalization.}
As personalized LLMs become increasingly complex, there is a growing need for systematic evaluation benchmarks. Existing work can be broadly categorized along two dimensions: the form of user-specific information and the evaluation setting. 
From the input perspective, prior benchmarks assess personalization using various forms of user information, including explicit profiles~\cite{zollo2024personalllm,zhao2025prefeval,tan2025personabench}, user-authored documents~\cite{salemi2023lamp,kumar2024longlamp}, and interaction histories with implicit or evolving preferences~\cite{maharana2024locomo,wu2024longmemeval,zhang2024memsim,jiang2025personamem,jiang2025personamem_v2,xiao2026alpsbench}. 
From the evaluation perspective, early benchmarks mainly consider relatively static personalization scenarios, where user information is explicitly provided or derived from a fixed set of documents or profile attributes~\cite{salemi2023lamp,kumar2024longlamp,zollo2024personalllm,zhao2025prefeval,tan2025personabench}. More recent efforts place greater emphasis on long-term memory and dynamic user modeling, evaluating whether models can retain user-related information across extended interactions, infer implicit preference signals from conversational histories, and adapt to preferences that evolve over time~\cite{maharana2024locomo,wu2024longmemeval,zhang2024memsim,jiang2025personamem,jiang2025personamem_v2,xiao2026alpsbench}.
However, these benchmarks remain largely confined to passive text-in-text-out settings, where personalization is evaluated primarily through generation rather than action, leaving a gap toward realistic assistant scenarios involving tool use and decision-making.

\paragraph{Benchmarks for LLM Agents.}

LLMs have evolved from text generators into autonomous agents capable of interacting with external tools and environments~\cite{schick2023toolformer}. Existing agent benchmarks have progressed from evaluating isolated tool-use capability to assessing increasingly realistic forms of interactive task execution.
Early benchmarks mainly focus on API invocation and tool-use accuracy, evaluating whether models can select appropriate tools and generate valid arguments for a given user request~\cite{li2023apibank,patil2024gorilla}. Subsequent benchmarks move toward more interactive and stateful settings, where agents must reason over multiple turns, track intermediate states, and respond to evolving context and feedback during execution~\cite{farn2023tooltalk,wang2024mint,qin2023toolllm,lu2024toolsandbox,shi2026aj}. More recent efforts further place agents in realistic execution environments, including web searching~\cite{liu2023agentbench,zhou2023webarena}, computer using~\cite{xie2024osworld}, software engineering~\cite{jimenez2024swebench}, and user-agent interaction~\cite{trivedi2024appworld,yao2024tau_bench,barres2025tau2_bench,he2025vitabench,li2026skillsbench,wang2026agentnoisebench,zheng2026risky}, to evaluate end-to-end task completion under real-world constraints.
However, existing agent benchmarks largely overlook personalization and typically assume that all task-relevant information is explicitly available in the current context, creating a gap with real-world assistant scenarios. Our work addresses this gap by jointly evaluating personalization and agentic execution in realistic interactive settings.
\section{VitaBench 2.0} 
\label{sec:methodology}

\begin{figure}[t]
    \centering
    \includegraphics[width=\linewidth]{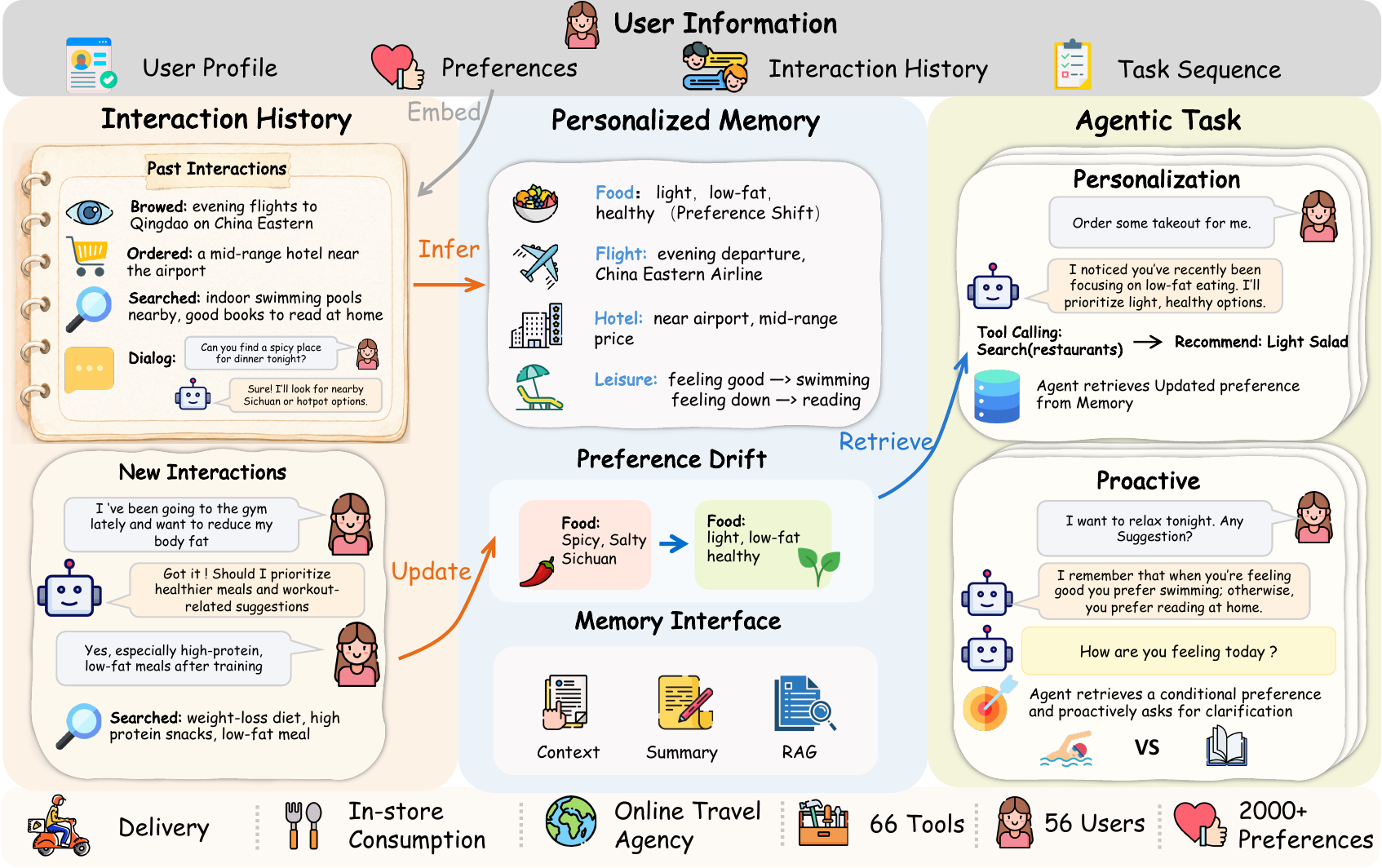}
\caption{Overview of VitaBench 2.0. The agents are required to operate over temporal task sequences for each user, infer evolving user preferences from fragmented interactions, maintain these preferences via a memory mechanism, and make personalized and proactive decisions.}
    \label{fig:overview}
\end{figure}

VitaBench 2.0 is designed to simulate long-term, user-agent collaboration scenarios for personalization and proactiveness evaluation, where agents are required to continuously satisfy user needs.
Figure~\ref{fig:overview} provides an overview. Each user $u$ is associated with a profile $P_u$, evolving preferences $\mathcal{N}_u = (n_1, n_2, \dots, n_L)$, and a temporal task sequence $\mathcal{T}_u = (t_1, t_2, \dots, t_N)$, designed to evaluate the agent’s ability to infer, maintain, and leverage user preferences over time.
Between tasks $t_{i-1}$ and $t_i$, the agent is exposed to newly introduced interaction histories that reflect emerging preferences or preference drift, and enabled to maintain a memory module $\mathcal{M}$ to store user information and support future decisions.
We describe the task formulation and the key modules in benchmark below.
We also provide a detailed analysis of our curated user profile and preferences in Appendix~\ref{app:benchmark_data_analysis}

\subsection{Task Set}

Tasks in VitaBench 2.0 are organized as temporally ordered sequences for individual users, where each sequence spans multiple domains.
Each individual task $t_i$ is an agentic task in which the agent interacts with domain-specific tools and an executable environment to fulfill a user request. Concretely, each task can be modeled as a partially observable Markov decision process (POMDP):
\begin{equation}
\mathcal{P}_i = (\mathcal{S}_i, \mathcal{A}_i, \mathcal{O}_i, \mathcal{T}_i, r_i),
\end{equation}
where $\mathcal{S}_i$ denotes the environment state, $\mathcal{A}_i$ the action space, $\mathcal{O}_i$ the observation space, $\mathcal{T}_i: \mathcal{S}_i \times \mathcal{A}_i \rightarrow \mathcal{S}_i$ the state transition function, and $r_i$ the task reward or evaluation function.

We design task complexity to arise from both tool-use and personalized user understanding, requiring agents to reason over explicit constraints from the user query and implicit signals derived from fragmented user interactions.
Specifically, a task instance is specified as:
\begin{equation}
t_i = (q_i, \mathcal{F}_i, \mathcal{E}_i, \mathcal{G}_i, \mathcal{H}_i),
\end{equation}
where $q_i$ is the user query, $\mathcal{F}_i$ is the set of available tools, $\mathcal{E}_i$ is the executable environment with underlying states, $\mathcal{G}_i$ is a set of evaluation rubrics, and $\mathcal{H}_i$ denotes the interaction histories exposed to agent between tasks $t_{i-1}$ and $t_i$, simulating fragmented user interactions over time.
Successful task execution requires the agent to identify user intent from $q_i$, select appropriate tools, and infer relevant user preferences from $\mathcal{H}_{1:i}$ to make consistent and personalized decisions.

Before solving task $t_i$, the agent is allow to updates its memory if enabled based on $\mathcal{H}_i$:
\begin{equation}
\mathcal{M}_i = \textsc{Update}(\mathcal{M}_{i-1}, \mathcal{H}_i).
\end{equation}
At each step $t$ within task $t_i$, the agent receives an observation $o_t \in \mathcal{O}_i$ consisting of the user query, dialogue history, and environment feedback from previous actions.
The agent then selects an action conditioned on the current observation and updated memory state:
\begin{equation}
a_t \sim \pi(a_t \mid o_t, \mathcal{M}_i),
\end{equation}
where the action space is given by
\begin{equation}
\mathcal{A}_i = \mathcal{A}_{\text{tool}} \cup \mathcal{A}_{\text{dialogue}},
\end{equation}
where $\mathcal{A}_{\text{tool}}$ denotes tool invocations and $\mathcal{A}_{\text{dialogue}}$ denotes natural-language responses to the user simulator.
After executing $a_t$, the environment transitions to a new state $s_{t+1}$ and returns a new observation $o_{t+1}$. The agent iterates between tool use and user interaction until the task is completed or a maximum number of steps is reached, producing a trajectory:
\begin{equation}
\tau_i = (o_0, a_0, o_1, a_1, \dots, o_T, a_T).
\end{equation}
The task accuracy is evaluated at both the trajectory level and outcome level by applying an evaluator LLM to $\tau_i$ and $a_T$ using the rubric set $\mathcal{G}_i$, which decomposes task success into a set of atomic criteria.
Inheriting from VitaBench~\cite{he2025vitabench}, we construct VitaBench 2.0 through systematic abstraction of real-world life-serving scenarios across three domains---Delivery, In-store Consumption, and Online Travel Agency---with a total of 66 tools.
Detailed descriptions of the task pipeline and environment construction are provided in Appendix~\ref{app:task} and Appendix~\ref{app:environment}.

\subsection{Key Module}

VitaBench 2.0 evaluates personalization by requiring agents to infer user preferences from fragmented historical interactions and leverage these preferences to collaborate with users. 
To support this evaluation, we carefully curate 56 users with more than 2,000 fine-grained preferences, covering diverse preference types and interaction contexts.
The construction of user profiles and preference distributions is data-driven, drawing inspiration from real-world user scenarios to better reflect realistic preference diversity and behavioral heterogeneity.
To reflect realistic long-term interaction scenarios, we allow the agent to maintain an external memory module that stores and updates user-specific information over time.
We next describe the construction of user profiles, user preferences, interaction histories, and the memory interface in detail.

\paragraph{User Profiles.}
Each user $u$ is associated with a manually curated profile $P_u$, constructed in a data-driven manner to reflect realistic user characteristics. 
To ensure both diversity and realism in the user population, we model users along multiple dimensions, including demographics, geographic and socioeconomic attributes, occupation, and social context, with distributions aligned to real-world scenario statistics. 
A comprehensive analysis of the curated profiles is provided in Appendix~\ref{app:user_profile_analysis}.

\paragraph{User Preferences.}
Each user $u$ is also associated with a set of preferences $\mathcal{N}_u = \{n_1, n_2, \dots, n_L\}$, spanning multiple aspects of daily life (e.g., dining, leisure and entertainment, shopping, travel, hobbies, and lifestyle habits).
Preferences are expressed as natural language statements grounded in the user profile (e.g., ``avoids spicy food due to a stomach condition'').
User preferences in real life are inherently dynamic. To simulate realistic evolution, we introduce temporally grounded preference drift events throughout each user’s task sequence. Between selected consecutive tasks, a subset of preferences may undergo one of three changes: (1)~\textit{addition}, where a new preference emerges; (2)~\textit{deletion}, where an existing preference becomes inactive; and (3)~\textit{modification}, where an existing preference shifts. 
In total, we manually curate 56 users with over 2,000 unique preferences. Detailed descriptions and illustrative examples are provided in Appendix~\ref{app:user}.
A comprehensive analysis of the curated preference is provided in Appendix~\ref{app:user_preference_analysis}.

\paragraph{Interaction History.}
User preferences are not explicitly provided to the agent, but are instead encoded in fragmented interaction histories accumulated over time. 
As the agent progresses from task $t_{i-1}$ to $t_i$, it is exposed to newly introduced interaction histories $\mathcal{H}_i$, which may reflect changes in the user’s underlying preferences.
Inspired by information accessibility in real-world scenarios, $\mathcal{H}_i$ contains two types of records: (1)~\textit{dialogues}, consisting of multi-turn user--agent conversations; and (2)~\textit{behaviors}, consisting of user behavior logs such as browsing, ordering, reviewing, and searching histories. 
Among these, not all interactions are preference-relevant. Instead, $\mathcal{H}_i$ can be viewed as comprising both signal interactions that reflect the user’s underlying preferences and noise interactions that are irrelevant, ambiguous, or contextually misleading.
This requires agents to distinguish consistent user preferences from irrelevant actions. 
Detailed construction of interaction history and illustrative examples are provided in Appendix~\ref{app:user}.

\paragraph{Memory Module.}
To capture long-term user dynamics across temporal task sequences, we allow agents to maintain an external memory module $\mathcal{M}$ for each user as a persistent representation of user-specific information. 
When memory is enabled, the agent interacts only with the memory module and does not have direct access to the full interaction histories.
Formally, before executing each task $t_i$, the agent is exposed to any newly available interaction history $H_i$ and updates its memory:
\begin{equation}
\mathcal{M}_i = \textsc{Update}(\mathcal{M}_{i-1}, H_i).
\end{equation}
During task execution, the agent conditions its actions on both the current observation and memory:
\begin{equation}
a_t \sim \pi(a_t \mid \textsc{Retrieve}(\mathcal{M}_i, q_i), o_t),
\end{equation}
where $\textsc{Retrieve}(\mathcal{M}_i, q_i)$ returns task-relevant information from memory.

To systematically study the role of memory in personalization, VitaBench 2.0 defines an extensible memory interface through two operations---$\textsc{Update}$ and $\textsc{Retrieve}$---allowing different memory architectures to be plugged in. 
Also, we implement two representative memory mechanisms:
\begin{itemize}[leftmargin=*]

\item \textbf{Agentic Memory.}
The agent maintains a structured representation of user information and actively controls the memory content by deciding what information to retain, update, or discard.
The memory is incrementally updated with each new history batch, and $\textsc{Retrieve}$ returns all or a selective memory representation.
This design requires the agent to perform selective abstraction, resolve conflicts across observations, and maintain long-term consistency.

\item \textbf{RAG Memory.}
Interaction records are stored in a memory bank with vector embeddings. 
$\textsc{Update}$ indexes new records, and $\textsc{Retrieve}$ performs similarity-based retrieval given the task query.
This design follows a fixed pipeline, where memory access is determined by retrieval without explicit control over what information is retained or discarded.

\end{itemize}
We provide a detailed discussion of memory mechanisms for agent systems in Appendix~\ref{app:memory}.

\subsection{Proactiveness}
\label{sec:proactiveness}

Beyond leveraging stored user preferences, an effective personalized agent should also know when its current knowledge is insufficient and proactively seek user clarification or conduct environment exploration. 
We evaluate this capability through proactive tasks, where successful task completion depends not only on retrieving the relevant user preference, but also on recognizing missing contextual information that cannot be inferred from memory or the current query alone.
Building on this idea, proactive tasks are constructed around missing but necessary information, where the correct action depends on contextual factors that are not directly observable to the agent. Solving such tasks requires the agent to capture the relevant conditional preference, recognize the unresolved ambiguity, and query the user or explore the environment before acting. These tasks are interleaved with standard personalization tasks, requiring the agent to adaptively decide when additional context is necessary rather than making decisions under incomplete information.

\section{Experiment}
\label{experiment}

\begin{table*}[!t]
    \setlength{\tabcolsep}{6pt}
    \caption{Performance of non-thinking and thinking models under different memory settings. The leaderboard is sorted by the Avg@4 on \textit{Full Context} setting. The best performance is in bold.}
    \centering
    \resizebox{\linewidth}{!}{%
    \begin{tabular}{@{\hspace{6pt}}l ccc ccc ccc@{\hspace{6pt}}}
    \toprule[1.2pt]
    \multirow{2}{*}{\textbf{Models}}
    & \multicolumn{3}{c@{\hspace{10pt}}}{\textbf{Full Context}}
    & \multicolumn{3}{c@{\hspace{10pt}}}{\textbf{Agentic Memory}}
    & \multicolumn{3}{c}{\textbf{RAG Memory}} \\
    \cmidrule(l{-2pt}r{6pt}){2-4}
    \cmidrule(l{6pt}r{6pt}){5-7}
    \cmidrule(l{6pt}r{-2pt}){8-10}
    & Avg@$4$ & Pass@$4$ & Pass\^{}$4$
    & Avg@$4$ & Pass@$4$ & Pass\^{}$4$
    & Avg@$4$ & Pass@$4$ & Pass\^{}$4$ \\
    \midrule

\rowcolor{gray!15}
\multicolumn{10}{c}{\textit{\textbf{Non-thinking Models}}} \\
\midrule

GPT-4o-mini (w/o thinking)               & 0.067 & 0.180 & 0.006 & 0.084 & 0.229 & 0.008 & 0.094 & 0.227 & 0.011 \\
GPT-3.5-Turbo (w/o thinking)             & 0.140 & 0.314 & 0.019 & 0.231 & 0.467 & 0.056 & 0.205 & 0.409 & 0.059 \\
LongCat-Flash-Chat (w/o thinking)        & 0.298 & 0.510 & 0.123 & 0.302 & 0.537 & 0.105 & 0.290 & 0.471 & 0.136 \\
GLM-4.5 (w/o thinking)                   & 0.307 & 0.529 & 0.127 & 0.330 & 0.569 & 0.112 & 0.316 & 0.523 & 0.152 \\
Doubao-Seed-1.6 (w/o thinking)           & 0.326 & 0.512 & 0.171 & 0.340 & 0.576 & 0.129 & 0.351 & 0.543 & 0.174 \\
GLM-4.6 (w/o thinking)                   & 0.342 & 0.612 & 0.113 & 0.336 & 0.623 & 0.084 & 0.317 & 0.555 & 0.123 \\
Kimi-K2.6 (w/o thinking)                 & 0.378 & 0.632 & 0.147 & 0.397 & 0.674 & 0.145 & 0.383 & 0.621 & 0.163 \\
GLM-5.1 (w/o thinking)                   & 0.420 & 0.654 & 0.204 & 0.423 & 0.664 & 0.182 & 0.383 & 0.585 & 0.200 \\
Doubao-Seed-2.0-pro (w/o thinking)       & 0.428 & 0.649 & 0.218 & 0.426 & 0.665 & 0.198 & 0.406 & 0.625 & 0.208 \\
DeepSeek-V4-Pro (w/o thinking)           & 0.456 & 0.652 & 0.267 & 0.427 & 0.658 & 0.207 & 0.424 & 0.618 & 0.247 \\

    \midrule
\rowcolor{gray!15}
\multicolumn{10}{c}{\textit{\textbf{Thinking Models}}} \\
\midrule

o4-mini (w/ thinking)                    & 0.210 & 0.433 & 0.047 & 0.270 & 0.533 & 0.073 & 0.261 & 0.452 & 0.091 \\
Gemini-2.5-Flash (w/ thinking)           & 0.282 & 0.556 & 0.063 & 0.312 & 0.567 & 0.098 & 0.309 & 0.544 & 0.107 \\
Qwen3-Max (w/ thinking)                  & 0.284 & 0.499 & 0.105 & 0.324 & 0.599 & 0.091 & 0.315 & 0.519 & 0.134 \\
Kimi-K2.6 (w/ thinking)                  & 0.293 & 0.533 & 0.099 & 0.280 & 0.508 & 0.088 & 0.303 & 0.511 & 0.118 \\
Gemini-2.5-Pro (w/ thinking)             & 0.331 & 0.605 & 0.109 & 0.378 & 0.638 & 0.138 & 0.320 & 0.579 & 0.109 \\
MiniMax-M2.7 (w/ thinking)               & 0.345 & 0.584 & 0.145 & 0.351 & 0.609 & 0.124 & 0.314 & 0.518 & 0.143 \\
GLM-4.6 (w/ thinking)                    & 0.359 & 0.612 & 0.116 & 0.351 & 0.625 & 0.107 & 0.336 & 0.574 & 0.135 \\
GLM-4.5 (w/ thinking)                    & 0.364 & 0.623 & 0.156 & 0.311 & 0.596 & 0.106 & 0.336 & 0.555 & 0.147 \\
Doubao-Seed-1.6 (w/ thinking)            & 0.373 & 0.599 & 0.176 & 0.383 & 0.646 & 0.123 & 0.375 & 0.591 & 0.179 \\
GLM-5.1 (w/ thinking)                    & 0.394 & 0.587 & 0.213 & 0.352 & 0.556 & 0.150 & 0.328 & 0.485 & 0.185 \\
DeepSeek-R1-0528 (w/ thinking)           & 0.396 & \textbf{0.691} & 0.131 & 0.412 & \textbf{0.712} & 0.118 & 0.390 & \textbf{0.643} & 0.153 \\
o3 (w/ thinking)                         & 0.403 & 0.653 & 0.169 & 0.401 & 0.669 & 0.154 & 0.362 & 0.587 & 0.158 \\
Claude-4.5-Sonnet (w/ thinking)          & 0.417 & 0.658 & 0.197 & 0.397 & 0.642 & 0.178 & 0.374 & 0.573 & 0.186 \\
GPT-5 (w/ thinking)                      & 0.441 & 0.658 & 0.226 & 0.421 & 0.647 & 0.204 & 0.410 & 0.591 & 0.236 \\
DeepSeek-V4-Pro (w/ thinking)            & 0.472 & 0.649 & 0.295 & 0.449 & 0.656 & 0.255 & \textbf{0.430} & 0.584 & 0.271 \\
Doubao-Seed-2.0-pro (w/ thinking)        & 0.474 & 0.683 & 0.270 & 0.428 & 0.650 & 0.225 & 0.339 & 0.496 & 0.205 \\
Claude-Opus-4.6 (w/ thinking)            & \textbf{0.503} & 0.664 & \textbf{0.337} & \textbf{0.454} & 0.645 & \textbf{0.259} & \textbf{0.430} & 0.566 & \textbf{0.299} \\

    \bottomrule[1.2pt]
    \end{tabular}}
    \label{tab:model_performance}
    \end{table*}

\subsection{Experimental Setups} \label{sec:setup}

\paragraph{Models.}
We evaluate a diverse set of state-of-the-art proprietary and open LLMs, covering both non-thinking and thinking configurations when available. 
The evaluated models include OpenAI family, including GPT-3.5-Turbo, GPT-4o-mini, GPT-5, and o-series models such as o3 and o4-mini~\citep{gpt41,gpt5,gpt51,gpt52,lrm-openaio3}; 
the DeepSeek family, including DeepSeek-R1 and DeepSeek-V4 variants~\citep{guo2025deepseekr1,deepseekv4}; 
Anthropic's Claude series, including Claude Sonnet and Claude Opus variants~\citep{anthropic2025claude4,anthropic2025claude45,claudeopus46}; 
Google's Gemini series, including Gemini-2.5-Flash and Gemini-2.5-Pro~\citep{comanici2025gemini25,gemini25pro,gemini25flash}; 
Qwen3-Max~\citep{qwen3max}; 
GLM variants, including GLM-4.5, GLM-4.6, and GLM-5.1~\citep{zeng2025glm45,zeng2025glm46,glm51}; 
ByteDance Seed series, including Seed-1.6 and Seed-2.0-Pro~\citep{doubao2025seed16,doubao2025seed20}; 
Kimi-K2.6~\citep{kimi26}; 
LongCat-Flash~\citep{team2025longcatflash}; 
and MiniMax-M2.7~\citep{minimax27}.
To ensure fair comparison, we distinguish between reasoning-enhanced (thinking) and non-reasoning (non-thinking) models. For hybrid architectures that support both modes, we evaluate the think-on and think-off configurations separately. 
We exclude smaller models due to the difficulty of the benchmark. The leaderboard is accordingly divided into thinking and non-thinking categories.

\paragraph{Implementations.}
All agents are implemented as function-calling agents based on the OpenAI tool schema. Interactions proceed without a predefined step limit and terminate either when the agent emits the token ``\#\#\#STOP\#\#\#'' or upon failure.
We use \texttt{gpt-4.1-2025-04-14} as the user simulator and evaluator.
Each task is run four times with a temperature of 0.0 for deterministic evaluation. 
Prompt templates for all components are detailed in Appendix~\ref{app:task}.
For memory mechanisms, we compare MemAgent~\cite{memagent} as agentic memory and the traditional RAG system as agent RAG memory.
For detailed implementation and configuration settings, please refer to Appendix~\ref{app:implementation_details}.

\paragraph{Metrics.}
We report Avg@4, Pass@4, and Pass\^{}$4$, computed from four independent runs and averaged over all tasks. 
Avg@4 measures the mean performance across the four runs. 
Pass@4 denotes the probability that at least one of the $4$ i.i.d.\ trials successfully completes the task. 
Pass\^{}$4$ represents the probability that all $4$ i.i.d.\ trials are successful.

\begin{figure}[!t]
    \centering
    \begin{minipage}{0.48\textwidth}
        \centering
        \includegraphics[width=\textwidth]{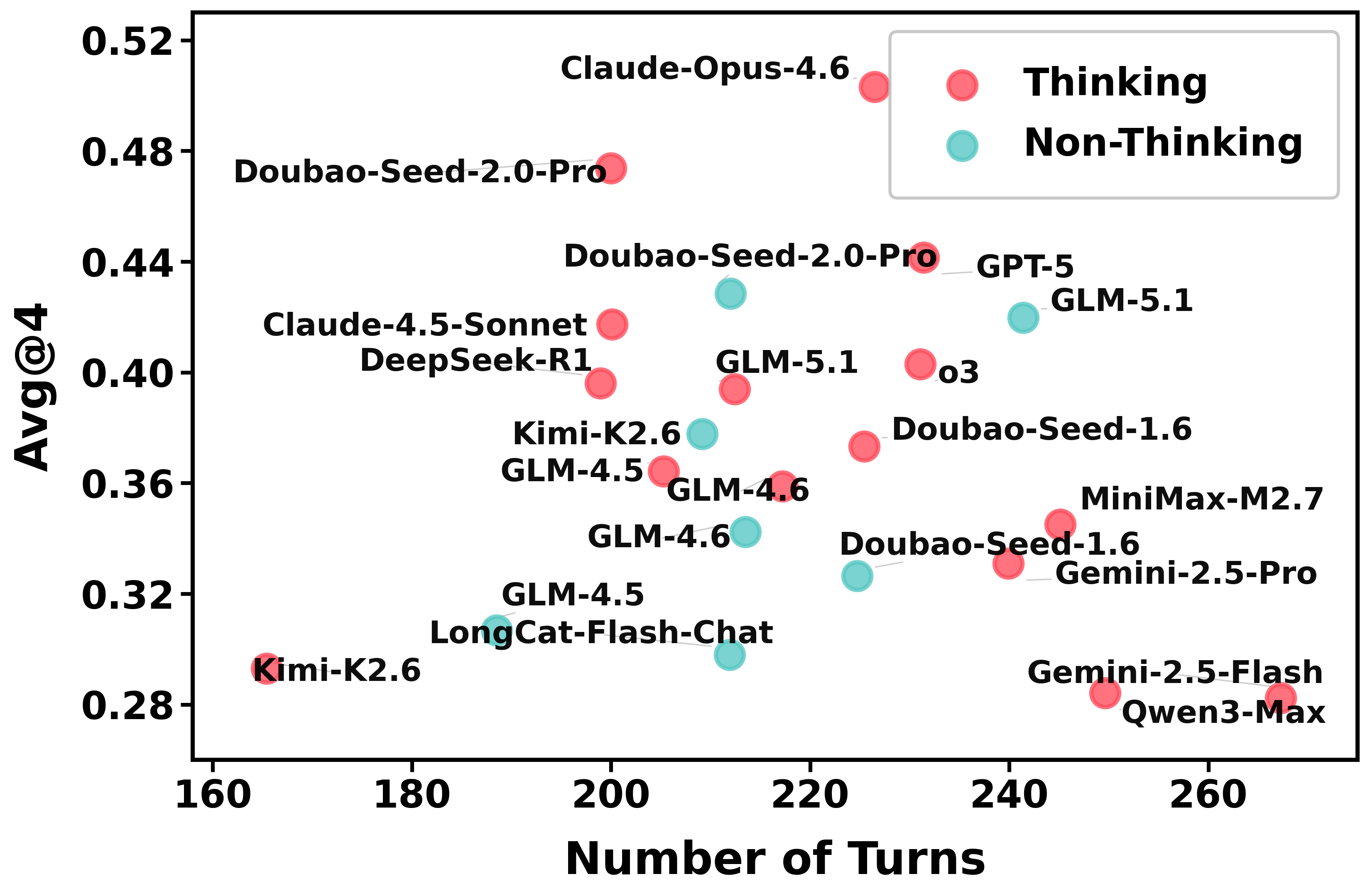}
        \caption{Average performance versus number of turns across models under full-context setting.}
        \label{fig:performance_turn}
    \end{minipage}
    \hfill
    \begin{minipage}{0.46\textwidth}
        \centering
        \includegraphics[width=\textwidth]{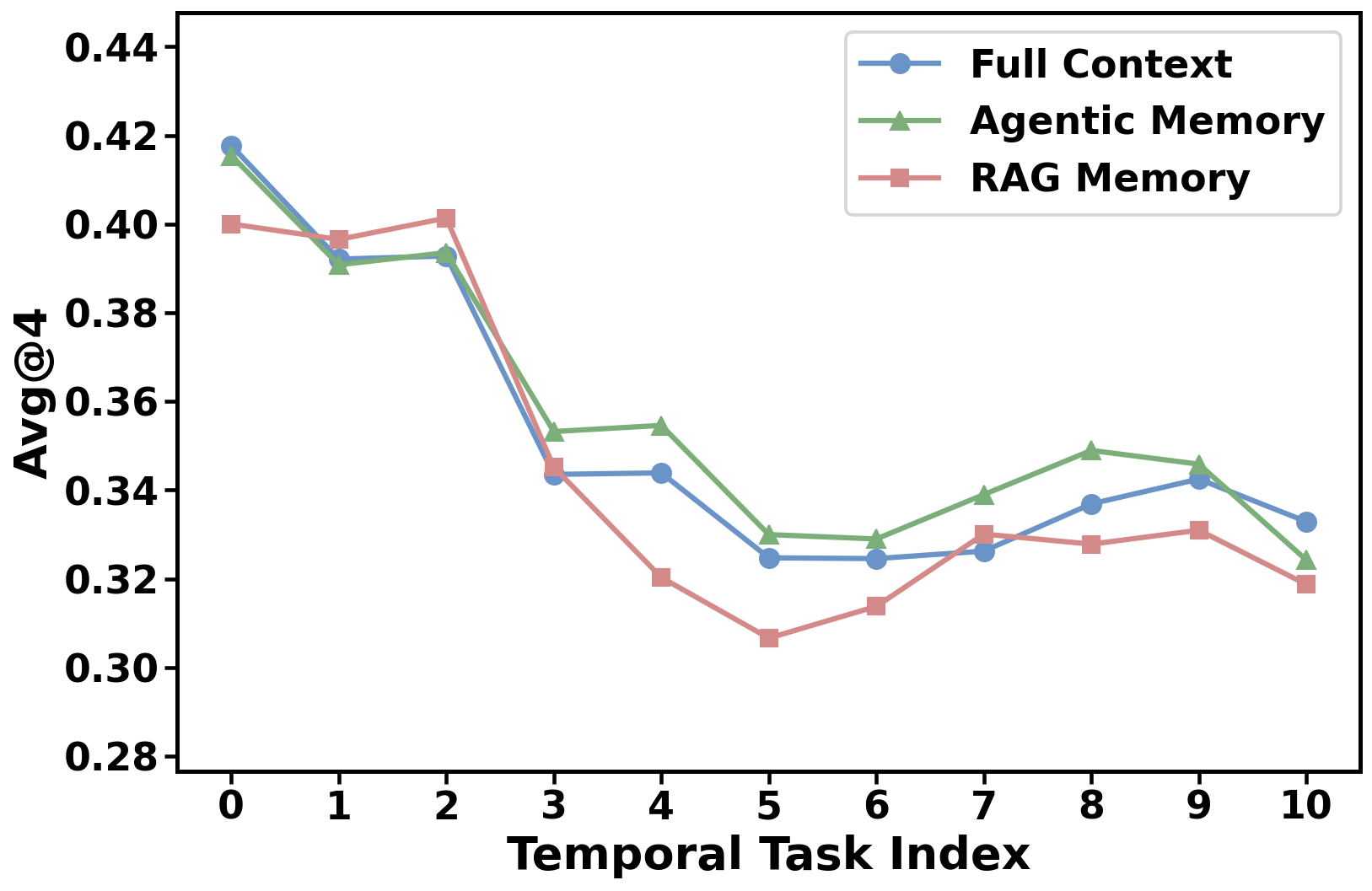}
        \caption{Average performance across tasks at each temporal task index.}
        \label{fig:subtask_reward}
    \end{minipage}
\end{figure}

\subsection{Main Results}

\begin{figure}[!t]
    \centering
    \includegraphics[width=\linewidth]{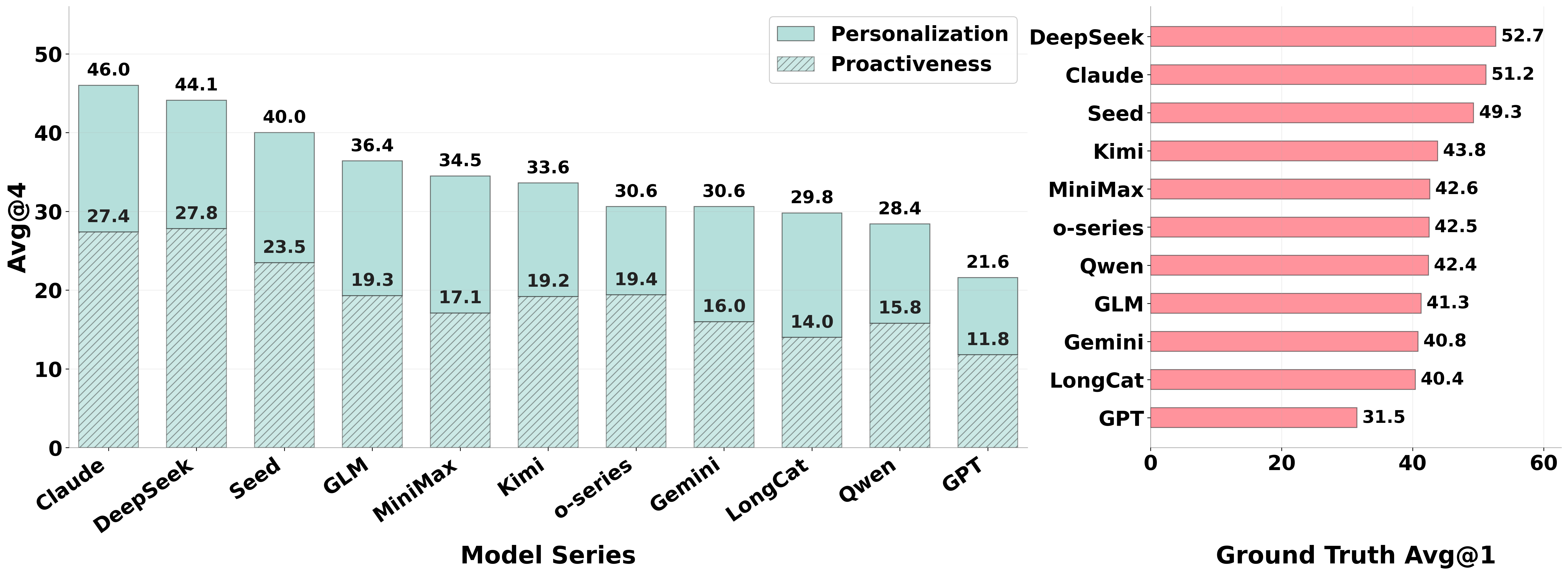}
\caption{Analysis of model behavior on VitaBench~2.0. \textbf{Left}: average performance on proactive tasks across model series. \textbf{Right}: performance on VitaBench~2.0 given ground-truth user preferences.}
    \label{fig:proactive}
\end{figure}

Table~\ref{tab:model_performance} presents evaluation results on VitaBench~2.0. We have the following observations.

\paragraph{Real-world personalization tasks remain highly challenging for current agents.}
Even under the \textit{Full Context} setting where full interaction history is accessible, state-of-the-art models achieve only Avg@$4$ of around $0.5$ and Pass\^{}$4$ of around $0.3$. 
This indicates that current agents struggle to reliably infer and utilize user preferences, despite already simplified task settings where tool complexity and instruction difficulty are intentionally controlled. 
Compared to traditional reasoning-intensive domains such as coding or mathematics, improvements from stronger base models are noticeably less pronounced, suggesting that personalization has emerged as a new bottleneck. 
As LLM agents are increasingly deployed in real-world user-facing applications, this gap highlights a fundamental limitation in their ability to support personalized decision-making.

\paragraph{Memory mechanisms play a critical but under-explored role.}
In realistic scenarios, interactions are often long-term and fragmented across sessions, making memory mechanisms essential for maintaining user representations. 
However, we observe that most models experience performance degradation when relying on memory, compared to the \textit{Full Context} setting. 
This trend holds for both agentic memory (where the model decides what to store and retrieve) and pipeline-based RAG memory. 
These results suggest that current agents are not yet capable of effectively utilizing memory, and that memory design remains a key challenge for improving long-term personalization.

\paragraph{Reasoning improvements do not directly translate to personalization gains.}
Unlike tasks that primarily depend on multi-step reasoning, enabling ``thinking'' modes does not consistently lead to better performance on VitaBench~2.0. 
While some models benefit from reasoning enhancements, the overall gains are modest and inconsistent across settings. 
As illustrated in Figure~\ref{fig:performance_turn}, enabling thinking mode does not consistently yield higher effectiveness (Avg@$4$) nor improved efficiency (\textit{Number of Turns}). 
This suggests that personalization requires capabilities beyond general reasoning, including robust preference extraction, long-term consistency, and the ability to handle noisy and incomplete observations. 
Consequently, advances in reasoning alone are insufficient to address the challenges of real-world personalized decision-making.

\begin{figure}[!t]
    \centering
    \includegraphics[width=0.9\linewidth]{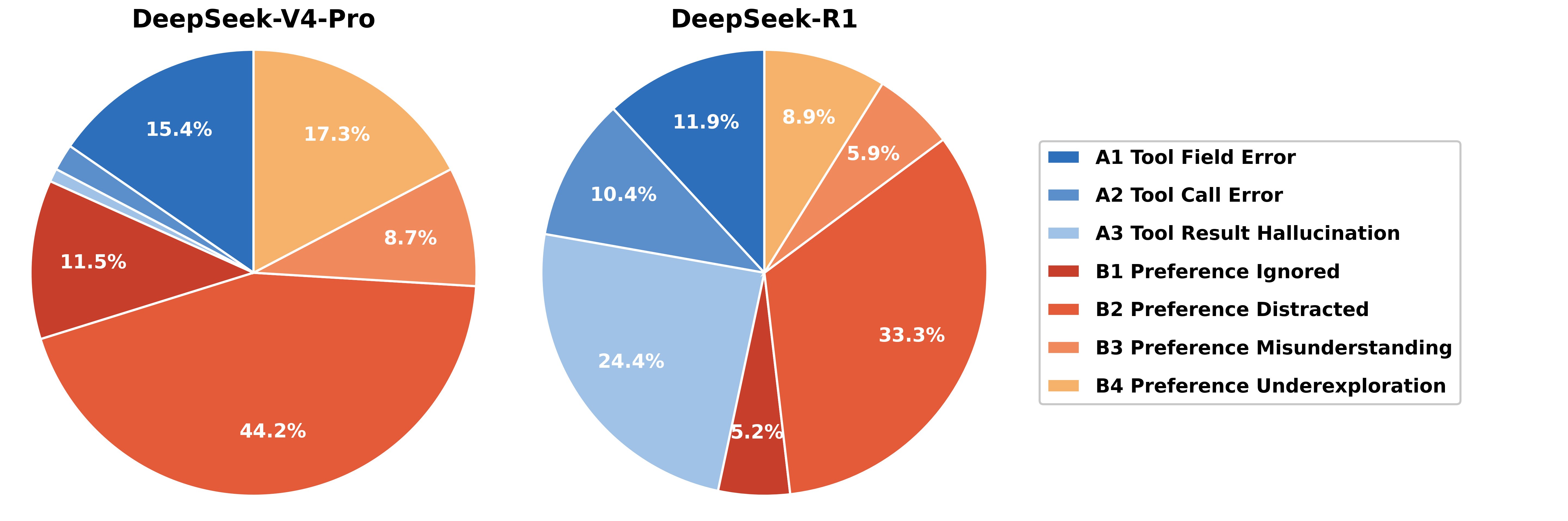}
\caption{Failure pattern statistics for DeepSeek-V4-Pro and DeepSeek-R1. Category A denotes tool-related errors, category B denotes preference-related errors.}
    \label{fig:filure_pattern}
\end{figure}

\subsection{Analysis \& Discussion}

\paragraph{Accumulated long-term user interactions pose a fundamental challenge for context handling and memory management.}
In VitaBench~2.0, tasks are organized as temporally ordered sequences for each user, spanning multiple domains. As the sequence progresses, interactions accumulate, leading to increasingly long and complex contexts for later tasks. To study this effect, we analyze the average performance at each temporal index across all users, reporting mean Avg@4 over all evaluated models. We report the task index of up to 10, as users have at least 10 tasks in their task sequences.
As shown in Figure~\ref{fig:subtask_reward}, performance generally degrades with increasing task index in both settings. Under full-context, this indicates a limitation in handling long interaction histories, where agents struggle to extract relevant signals from long context. Under memory-based settings, the degradation is further amplified by imperfect memory management: repeated $\textsc{Update}$ and $\textsc{Retrieve}$ operations introduce information loss and error accumulation, causing early inaccuracies to propagate to later tasks. These results highlight that both long-context reasoning and effective memory utilization remain key bottlenecks for current agent systems.

\paragraph{Current agents struggle to recognize missing information and engage in proactive interactions with users.}
We evaluate proactive capabilities by measuring performance on tasks that require agents to actively identify missing information and query users or explore the environment before making decisions. As shown in Figure~\ref{fig:proactive} (left), proactive performance is consistently lower than personalization performance across all model families. For example, while Claude achieves an average personalization score of 46.0, its proactive score drops to 27.4; similar gaps are observed for DeepSeek (44.1 vs.\ 27.8) and GLM (36.4 vs.\ 19.3).
These results indicate that current agents often fail to recognize when their knowledge is insufficient, and instead proceed with incomplete information rather than initiating clarification. This limitation suggests that proactive interaction remains underdeveloped in current agent systems.

\paragraph{Even given ground-truth user preferences, effectively leveraging them remains challenging for current agents.}
To isolate the difficulty of preference utilization, we provide models with ground-truth user preferences for each task and evaluate their performance. As shown in Figure~\ref{fig:proactive} (right), performance improves compared to the standard setting but remains far from optimal. For instance, DeepSeek and Claude achieve 52.7 and 51.2 under ground-truth preferences, while most other model families remain below 50 (e.g., Seed 49.3, Kimi 43.8, MiniMax 42.6).
These results suggest that the challenge of personalization is not solely due to preference extraction and maintain, but also arises from difficulties in reasoning over, prioritizing, and consistently applying preference information during decision-making. Even when provided with accurate user profiles, current agents often fail to translate this information into effective actions.

\paragraph{Failure pattern analysis shows that personalization emerges as the primary bottleneck in agent performance.}
We analyze failure patterns of agents on VitaBench~2.0. Specifically, we use Claude-Opus-4.6 as an external analyzer to examine the full trajectories of DeepSeek family models, including DeepSeek-V4-Pro and DeepSeek-R1, and categorize their errors into fine-grained types. The results are summarized in Figure~\ref{fig:filure_pattern}.
We observe that the majority of errors in VitaBench~2.0 stem from failures in capturing and utilizing user preferences. In many cases, even state-of-the-art agents fail to correctly infer user preferences from historical interactions, or neglect them during decision-making. For example, agents often default to selecting high-rated or popular items, instead of aligning with user-specific preferences inferred from prior behavior.
Moreover, we observe a shift in failure patterns across model generations. Earlier models tend to suffer more from tool-related errors due to limitations in their base capabilities, whereas stronger models exhibit fewer tool failures but still struggle with personalization, making it the dominant bottleneck. This suggests that, as foundational reasoning and tool-use abilities improve, personalization becomes the next critical challenge for advancing agent performance.

\section{Limitation} 
\label{sec:limitation}
While VitaBench 2.0 provides a principled benchmark for studying personalization and proactiveness in LLM agents, it has several limitations.
First, user preferences and interaction histories are programmatically constructed to allow precise control over preference dynamics and task difficulty. While this facilitates reproducible evaluation, it may not fully capture the full diversity of real-world user behavior.
Second, the memory interface abstracts memory into update and retrieval operations, enabling controlled comparison across different designs. This abstraction focuses on isolating the role of memory, and does not aim to cover all possible end-to-end architectures.
Third, evaluation is based on rubric-driven assessment over task trajectories, providing structured and interpretable signals. More open-ended measures of user satisfaction are beyond the scope of this work.
Overall, these design choices are intended to prioritize controllability and comparability, and we view VitaBench 2.0 as a complementary testbed for studying core challenges in personalized and proactive agents.
\section{Conclusion}
\label{sec:conclusion}

In this work, we introduce VitaBench 2.0, a benchmark for evaluating personalization and proactiveness in LLM-based agents. 
VitaBench 2.0 organizes tasks as user-centric sequences, embeds evolving preferences into fragmented interaction histories, and incrementally exposes these histories to the agent, capturing the key challenge of inferring and updating user preferences over time.
To support systematic analysis, we design an extensible memory interface that enables controlled comparison of different memory mechanisms within a unified framework. 
Through extensive experiments on a diverse set of frontier models, we find that current LLM agents struggle to reliably infer, utilize, and update user preferences, especially when preferences evolve or when information is incomplete. Further analysis provides insights into the failure modes of current agents and the difficulties of long-term preference modeling.
Overall, VitaBench 2.0 reveals a significant gap between existing LLM agents and realistic personalized assistants, and provides a testbed for advancing research on memory, personalization, and proactive behavior.


\bibliographystyle{unsrtnat}
\bibliography{neurips_2024}

@article{liu2025survey_personalized_llm,
  title={A Survey of Personalized Large Language Models: Progress and Future Directions},
  author={Liu, Jiahong and Qiu, Zexuan and Li, Zhongyang and Dai, Quanyu and Yu, Wenhao and Zhu, Jieming and Hu, Minda and Yang, Menglin and Chua, Tat-Seng and King, Irwin},
  journal={arXiv preprint arXiv:2502.11528},
  year={2025}
}

@article{zhang2024personalization_survey,
  title={Personalization of Large Language Models: A Survey},
  author={Zhang, Zhehao and Lutz, Ryan and Mao, Aidan and Bao, Tianyue and Wang, Zijian and Zhao, Zhoujian and Xiang, Kaixin and Ding, Liwei and Tong, Le and Zhuo, Jiaxin and others},
  journal={arXiv preprint arXiv:2411.00027},
  year={2024}
}

@article{tseng2024two_tales_persona,
  title={Two Tales of Persona in {LLM}s: A Survey of Role-Playing and Personalization},
  author={Tseng, Yu-Min and Huang, Yu-Chao and Hsiao, Teng-Yun and Huang, Wei-Lin and others},
  journal={arXiv preprint arXiv:2406.01171},
  year={2024}
}

@inproceedings{salemi2024optimization_ropg,
  title={Optimization Methods for Personalizing Large Language Models through Retrieval Augmentation},
  author={Salemi, Alireza and Mysore, Sheshera and Bendersky, Michael and Zamani, Hamed},
  booktitle={Proceedings of the 47th International ACM SIGIR Conference on Research and Development in Information Retrieval},
  year={2024}
}

@inproceedings{mysore2023pearl,
  title={{PEARL}: Personalizing Large Language Model Writing Assistants with Generation-Calibrated Retrievers},
  author={Mysore, Sheshera and Lu, Zhuoran and Wan, Mengting and McAuley, Julian and Zamani, Hamed},
  booktitle={Proceedings of the 1st Workshop on Customizable NLP},
  year={2024}
}

@article{richardson2023pag,
  title={Integrating Summarization and Retrieval for Enhanced Personalization via Large Language Models},
  author={Richardson, Jesse and Bloom, Kristen and Founta, Aggeliki and Mathew, Brendan},
  journal={arXiv preprint arXiv:2310.20081},
  year={2023}
}

@article{li2023teach_llm_personalize,
  title={Teach {LLM}s to Personalize--An Approach Inspired by Writing Education},
  author={Li, Cheng and Chen, Mingyang and Wang, Haoping and Zhu, Bin and Luo, Haoyu and others},
  journal={arXiv preprint arXiv:2308.07968},
  year={2023}
}

@article{wu2024understanding_user_profile,
  title={Understanding the Role of User Profile in the Personalization of Large Language Models},
  author={Wu, Ostap and Haim, Max and Dey, Tanmay and others},
  journal={arXiv preprint arXiv:2406.17803},
  year={2024}
}

@inproceedings{tan2024oppu,
  title={Democratizing Large Language Models via Personalized Parameter-Efficient Fine-Tuning},
  author={Tan, Zhaoxuan and Zeng, Qingkai and Tian, Yijun and Liu, Zheyuan and Yin, Bing and Jiang, Meng},
  booktitle={Proceedings of the 2024 Conference on Empirical Methods in Natural Language Processing},
  year={2024}
}

@inproceedings{zhang2024plora,
  title={{PLoRA}: Personalized Low-Rank Adaptation for Human-Centered Text Understanding},
  author={Zhang, Yuting and Ding, Yuliang and others},
  booktitle={Proceedings of the AAAI Conference on Artificial Intelligence},
  year={2024}
}

@article{zhuang2024hydra,
  title={{HYDRA}: Model Factorization Framework for Black-Box {LLM} Personalization},
  author={Zhuang, Tao and Wang, Xin and Yuan, Zhirui and others},
  journal={arXiv preprint arXiv:2406.02888},
  year={2024}
}

@article{jang2023personalized_soups,
  title={Personalized Soups: Personalized Large Language Model Alignment via Post-hoc Parameter Merging},
  author={Jang, Joel and Kim, Seungone and Lin, Bill Yuchen and Wang, Yizhong and Shafran, Jack and Choi, Yejin and others},
  journal={arXiv preprint arXiv:2310.11564},
  year={2023}
}

@article{zhou2023modpo,
  title={Beyond One-Preference-Fits-All Alignment: Multi-Objective Direct Preference Optimization},
  author={Zhou, Zhanhui and Liu, Jie and Dong, Jing and Yang, Jiaheng and others},
  journal={arXiv preprint arXiv:2310.03708},
  year={2023}
}

@article{zhao2025nextquill,
  title={{NextQuill}: Causal Preference Modeling for Enhancing {LLM} Personalization},
  author={Zhao, Xiaoyan and You, Juntao and Zhang, Yang and Wang, Wenjie and Cheng, Hong and Feng, Fuli and Ng, See-Kiong and Chua, Tat-Seng},
  journal={arXiv preprint arXiv:2506.02368},
  year={2025}
}

@article{packer2023memgpt,
  title={{MemGPT}: Towards {LLM}s as Operating Systems},
  author={Packer, Charles and Wooders, Sarah and Lin, Kevin and Fang, Vivian and Patil, Shishir G and Stoica, Ion and Gonzalez, Joseph E},
  journal={arXiv preprint arXiv:2310.08560},
  year={2023}
}

@misc{mem0_2024,
  title={Mem0: The Memory Layer for Personalized {AI}},
  author={{Mem0}},
  howpublished={\url{https://mem0.ai}},
  year={2024}
}

@article{xu2025amem,
  title={{A-MEM}: Agentic Memory for {LLM} Agents},
  author={Xu, Wujiang and Liang, Zujie and Mei, Kai and others},
  journal={arXiv preprint arXiv:2502.12110},
  year={2025}
}

@inproceedings{salemi2023lamp,
  title={{LaMP}: When Large Language Models Meet Personalization},
  author={Salemi, Alireza and Mysore, Sheshera and Bendersky, Michael and Zamani, Hamed},
  booktitle={Proceedings of the 62nd Annual Meeting of the Association for Computational Linguistics},
  year={2024}
}

@article{kumar2024longlamp,
  title={{LongLaMP}: A Benchmark for Personalized Long-Form Text Generation},
  author={Kumar, Ishita and Viswanathan, Snigdha and others},
  journal={arXiv preprint arXiv:2407.11016},
  year={2024}
}

@article{zollo2024personalllm,
  title={{PersonalLLM}: Tailoring {LLM}s to Individual Preferences},
  author={Zollo, Thomas P and Weidinger, Andrew and others},
  journal={arXiv preprint arXiv:2409.20296},
  year={2024}
}

@inproceedings{zhao2025prefeval,
  title={Do {LLM}s Recognize Your Preferences? Evaluating Personalized Preference Following in {LLM}s},
  author={Zhao, Xiaoyan and Zhang, Yang and You, Juntao and Wang, Wenjie and Feng, Fuli and others},
  booktitle={International Conference on Learning Representations},
  year={2025}
}

@inproceedings{tan2025personabench,
  title={{PersonaBench}: Evaluating {AI} Models on Understanding Personal Information through Accessing (Synthetic) Private User Data},
  author={Tan, Zhaoxuan and others},
  booktitle={International Conference on Learning Representations},
  year={2025}
}

@inproceedings{maharana2024locomo,
  title={Evaluating Very Long-Term Conversational Memory of {LLM} Agents},
  author={Maharana, Adyasha and Lee, Dong-Ho and Tulyakov, Sergey and Bansal, Mohit and Barbieri, Francesco and Fang, Yuwei},
  booktitle={Proceedings of the 62nd Annual Meeting of the Association for Computational Linguistics},
  year={2024}
}

@article{wu2024longmemeval,
  title={{LongMemEval}: Benchmarking Chat Assistants on Long-Term Interactive Memory},
  author={Wu, Di and Wang, Hongwei and Yu, Wenhao and Wu, Yuwei and Yu, Kai and others},
  journal={arXiv preprint arXiv:2410.10813},
  year={2024}
}

@article{zhang2024memsim,
  title={{MemSim}: A {B}ayesian Simulator for Evaluating Memory of Personal Assistants},
  author={Zhang, Zeyu and others},
  journal={arXiv preprint arXiv:2409.20163},
  year={2024}
}

@article{jiang2025personamem,
  title={Know Me, Respond to Me: Benchmarking {LLM}s for Dynamic User Profiling and Personalized Responses at Scale},
  author={Jiang, Bowen and Hao, Zhuoqun and Cho, Young-Min and Li, Bryan and Yuan, Yuan and Chen, Sihao and Ungar, Lyle and Taylor, Camillo J and Roth, Dan},
  journal={arXiv preprint arXiv:2504.14225},
  year={2025}
}

@article{jiang2025personamem_v2,
  title={{PersonaMem-v2}: Towards Personalized Intelligence via Learning Implicit User Personas and Agentic Memory},
  author={Jiang, Bowen and Yuan, Yuan and Shen, Maohao and Hao, Zhuoqun and Xu, Zhangchen and Chen, Zichen and others},
  journal={arXiv preprint arXiv:2512.06688},
  year={2025}
}

@article{xiao2026alpsbench,
  title={{AlpsBench}: An {LLM} Personalization Benchmark for Real-Dialogue Memorization and Preference Alignment},
  author={Xiao, Jianfei and Yu, Xiang and Wang, Chengbing and Zheng, Wuqiang and Lin, Xinyu and Liu, Kaining and Ding, Hongxun and Zhang, Yang and Wang, Wenjie and Feng, Fuli and He, Xiangnan},
  journal={arXiv preprint arXiv:2603.26680},
  year={2026}
}

@inproceedings{schick2023toolformer,
  title={Toolformer: Language Models Can Teach Themselves to Use Tools},
  author={Schick, Timo and Dwivedi-Yu, Jane and Dess{\`\i}, Roberto and Raileanu, Roberta and Lomeli, Maria and Hambro, Eric and Zettlemoyer, Luke and Cancedda, Nicola and Scialom, Thomas},
  booktitle={Advances in Neural Information Processing Systems},
  year={2023}
}

@inproceedings{li2023apibank,
  title={{API-Bank}: A Comprehensive Benchmark for Tool-Augmented {LLM}s},
  author={Li, Minghao and Song, Feifan and Yu, Bowen and Yu, Haiyang and Li, Zhoujun and Huang, Fei and Li, Yongbin},
  booktitle={Proceedings of the 2023 Conference on Empirical Methods in Natural Language Processing},
  year={2023}
}

@inproceedings{patil2024gorilla,
  title={Gorilla: Large Language Model Connected with Massive {API}s},
  author={Patil, Shishir G and Zhang, Tianjun and Wang, Xin and Gonzalez, Joseph E},
  booktitle={Advances in Neural Information Processing Systems},
  year={2024}
}

@article{qin2023toolllm,
  title={{ToolLLM}: Facilitating Large Language Models to Master 16000+ Real-World {API}s},
  author={Qin, Yujia and Liang, Shihao and Ye, Yining and Zhu, Kunlun and Yan, Lan and Lu, Yaxi and Lin, Yankai and Cong, Xin and Tang, Xiangru and others},
  journal={arXiv preprint arXiv:2307.16789},
  year={2023}
}

@article{farn2023tooltalk,
  title={{ToolTalk}: Evaluating Tool-Usage in a Conversational Setting},
  author={Farn, Nicholas and Shin, Richard},
  journal={arXiv preprint arXiv:2311.10775},
  year={2023}
}

@inproceedings{wang2024mint,
  title={{MINT}: Evaluating {LLM}s in Multi-turn Interaction with Tools and Language Feedback},
  author={Wang, Xingyao and Wang, Zihan and Liu, Jiateng and Chen, Yangyi and Yuan, Lifan and Peng, Hao and Ji, Heng},
  booktitle={International Conference on Learning Representations},
  year={2024}
}

@article{lu2024toolsandbox,
  title={{ToolSandbox}: A Stateful, Conversational, Interactive Evaluation Benchmark for {LLM} Tool Use Capabilities},
  author={Lu, Jiarui and Zhu, Thomas and Jiang, Hao and Skreta, Marta and Rawat, Arun Sai and others},
  journal={arXiv preprint arXiv:2408.04682},
  year={2024}
}

@inproceedings{liu2023agentbench,
  title={{AgentBench}: Evaluating {LLM}s as Agents},
  author={Liu, Xiao and Yu, Hao and Zhang, Hanchen and Xu, Yifan and Lei, Xuanyu and Lai, Hanyu and Gu, Yu and others},
  booktitle={International Conference on Learning Representations},
  year={2024}
}

@inproceedings{zhou2023webarena,
  title={{WebArena}: A Realistic Web Environment for Building Autonomous Agents},
  author={Zhou, Shuyan and Xu, Frank F and Zhu, Hao and Zhou, Xuhui and Lo, Robert and Sridhar, Abishek and Cheng, Xianyi and Bisk, Yonatan and Fried, Daniel and Alon, Uri and others},
  booktitle={International Conference on Learning Representations},
  year={2024}
}

@inproceedings{xie2024osworld,
  title={{OSWorld}: Benchmarking Multimodal Agents for Open-Ended Tasks in Real Computer Environments},
  author={Xie, Tianbao and Zhang, Danyang and Chen, Jixuan and Li, Xiaochuan and Zhao, Siheng and Cao, Ruisheng and Hua, Toh Jing and others},
  booktitle={Advances in Neural Information Processing Systems},
  year={2024}
}

@inproceedings{trivedi2024appworld,
  title={{AppWorld}: A Controllable World of Apps and People for Benchmarking Interactive Coding Agents},
  author={Trivedi, Harsh and Khot, Tushar and Hartmann, Mareike and Manber, Reshef and Baber, Vinty and Fishi, David and others},
  booktitle={Proceedings of the 62nd Annual Meeting of the Association for Computational Linguistics},
  year={2024}
}

@article{yao2024tau_bench,
  title={$\tau$-bench: A Benchmark for Tool-Agent-User Interaction in Real-World Domains},
  author={Yao, Shunyu and Shinn, Noah and Razavi, Pedram and Narasimhan, Karthik},
  journal={arXiv preprint arXiv:2406.12045},
  year={2024}
}

@article{barres2025tau2_bench,
  title={$\tau^2$-Bench: Evaluating Conversational Agents in a Dual-Control Environment},
  author={Barres, Victor and Dong, Honghua and Ray, Soham and Si, Xujie and Narasimhan, Karthik},
  journal={arXiv preprint arXiv:2506.07982},
  year={2025}
}

@article{he2025vitabench,
  title={{VitaBench}: Benchmarking {LLM} Agents with Versatile Interactive Tasks in Real-world Applications},
  author={He, Wei and Sun, Yueqing and Hao, Hongyan and Hao, Xueyuan and Xia, Zhikang and Gu, Qi and Han, Chengcheng and others},
  journal={arXiv preprint arXiv:2509.26490},
  year={2025}
}

@article{li2026skillsbench,
  title={{SkillsBench}: Benchmarking How Well Agent Skills Work Across Diverse Tasks},
  author={Li, Xiangyi and Chen, Wenbo and Liu, Yimin and Zheng, Shenghan and others},
  journal={arXiv preprint arXiv:2602.12670},
  year={2026}
}

@inproceedings{jimenez2024swebench,
  title={{SWE-bench}: Can Language Models Resolve Real-World {GitHub} Issues?},
  author={Jimenez, Carlos E and Yang, John and Wettig, Alexander and Yao, Shunyu and Pei, Kexin and Press, Ofir and Narasimhan, Karthik},
  booktitle={International Conference on Learning Representations},
  year={2024}
}

@article{lrm-openaio3,
  author  = {{OpenAI}},
  title   = {Introducing o3 and o4-mini},
  year    = {2025},
  url     = {https://openai.com/index/introducing-o3-and-o4-mini/}
}

@article{gpt41,
  author  = {{OpenAI}},
  title   = {Introducing GPT-4.1 in the API},
  year    = {2025},
  url     = {https://openai.com/index/gpt-4-1/}
}

@article{gpt5,
  author  = {{OpenAI}},
  title   = {Introducing GPT-5},
  year    = {2025},
  url     = {https://openai.com/index/introducing-gpt-5/}
}

@article{gpt51,
  author  = {{OpenAI}},
  title   = {Introducing GPT-5.1},
  year    = {2025},
  url     = {https://openai.com/index/gpt-5-1/}
}

@article{gpt52,
  author  = {{OpenAI}},
  title   = {Introducing GPT-5.2},
  year    = {2025},
  url     = {https://openai.com/index/introducing-gpt-5-2/}
}

@article{guo2025deepseekr1,
  title   = {DeepSeek-R1: Incentivizing Reasoning Capability in LLMs via Reinforcement Learning},
  author  = {Guo, Daya and others},
  journal = {arXiv preprint arXiv:2501.12948},
  year    = {2025}
}

@article{deepseekv31,
  title   = {DeepSeek-V3.1 Model Card},
  author  = {{DeepSeekAI}},
  year    = {2025},
  url     = {https://huggingface.co/deepseek-ai/DeepSeek-V3.1}
}

@article{liu2025deepseekv32,
  title   = {DeepSeek-V3.2: Pushing the Frontier of Open Large Language Models},
  author  = {Liu, Aixin and others},
  journal = {arXiv preprint arXiv:2512.02556},
  year    = {2025}
}

@article{anthropic2025claude4,
  author  = {{Anthropic}},
  title   = {Claude Sonnet 4 System Card},
  year    = {2025},
  url     = {https://www.anthropic.com/news/claude-4}
}

@article{anthropic2025claude45,
  author  = {{Anthropic}},
  title   = {Claude Sonnet 4.5 Model Card},
  year    = {2025},
  url     = {https://www.anthropic.com/news/claude-sonnet-4-5}
}

@article{comanici2025gemini25,
  title   = {Gemini 2.5: Advanced Reasoning, Multimodality, and Agentic Capabilities},
  author  = {Comanici, Gheorghe and others},
  journal = {arXiv preprint arXiv:2507.06261},
  year    = {2025}
}

@article{gemini25pro,
  author  = {{Google}},
  title   = {Gemini 2.5 Pro Model Card},
  year    = {2025},
  url     = {https://modelcards.withgoogle.com/assets/documents/gemini-2.5-pro.pdf}
}

@article{gemini25flash,
  author  = {{Google}},
  title   = {Gemini 2.5 Flash Model Card},
  year    = {2025},
  url     = {https://storage.googleapis.com/deepmind-media/Model-Cards/Gemini-2-5-Flash-Model-Card.pdf}
}

@article{qwen3max,
  author  = {{Qwen Team}},
  title   = {Qwen3-Max Model Card},
  year    = {2025},
  url     = {https://qwen.ai/blog?id=qwen3-max}
}

@article{zeng2025glm45,
  title   = {GLM-4.5: Agentic, Reasoning, and Coding Foundation Models},
  author  = {Zeng, Aohan and others},
  journal = {arXiv preprint arXiv:2508.06471},
  year    = {2025}
}

@article{zeng2025glm46,
  author  = {{Z.ai}},
  title   = {GLM-4.6 Technical Blog},
  year    = {2025},
  url     = {https://z.ai/blog/glm-4.6}
}

@article{doubao2025seed16,
  author  = {{ByteDance}},
  title   = {Seed 1.6 Technical Introduction},
  year    = {2025},
  url     = {https://seed.bytedance.com/en/seed1_6}
}

@article{team2025longcatflash,
  title   = {LongCat-Flash Technical Report},
  author  = {{Meituan LongCat Team}},
  journal = {arXiv preprint arXiv:2509.01322},
  year    = {2025}
}

@article{longcatflash2601,
  author       = {Meituan LongCat Team},
  title        = {LongCat-Flash-Thinking-2601 Technical Report},
  journal      = {CoRR},
  volume       = {abs/2601.16725},
  year         = {2026}
}

@article{amem,
  author       = {Wujiang Xu and
                  Zujie Liang and
                  Kai Mei and
                  Hang Gao and
                  Juntao Tan and
                  Yongfeng Zhang},
  title        = {{A-MEM:} Agentic Memory for {LLM} Agents},
  journal      = {CoRR},
  volume       = {abs/2502.12110},
  year         = {2025}
}

@article{memagent,
  author       = {Hongli Yu and
                  Tinghong Chen and
                  Jiangtao Feng and
                  Jiangjie Chen and
                  Weinan Dai and
                  Qiying Yu and
                  Ya{-}Qin Zhang and
                  Wei{-}Ying Ma and
                  Jingjing Liu and
                  Mingxuan Wang and
                  Hao Zhou},
  title        = {MemAgent: Reshaping Long-Context {LLM} with Multi-Conv RL-based Memory
                  Agent},
  journal      = {CoRR},
  volume       = {abs/2507.02259},
  year         = {2025}
}

@inproceedings{xiao2024streamingllm,
  title     = {Efficient Streaming Language Models with Attention Sinks},
  author    = {Xiao, Guangxuan and Tian, Yuandong and Chen, Beidi and Han, Song and Lewis, Mike},
  booktitle = {International Conference on Learning Representations (ICLR)},
  year      = {2024}
}

@inproceedings{wang2023longmem,
  title     = {Augmenting Language Models with Long-Term Memory},
  author    = {Wang, Weizhi and Dong, Li and Cheng, Hao and Liu, Xiaodong and Yan, Xifeng and Gao, Jianfeng and Wei, Furu},
  booktitle = {Advances in Neural Information Processing Systems (NeurIPS)},
  year      = {2023}
}

@article{wang2023recursivesum,
  title   = {Recursively Summarizing Enables Long-Term Dialogue Memory in Large Language Models},
  author  = {Wang, Qingyue and Ding, Liang and Cao, Yanan and Tian, Zhiliang and Wang, Shi and Tao, Dacheng and Guo, Li},
  journal = {arXiv preprint arXiv:2308.15022},
  year    = {2023}
}

@inproceedings{zhang2024coa,
  title     = {Chain of Agents: Large Language Models Collaborating on Long-Context Tasks},
  author    = {Zhang, Yusen and Sun, Ruoxi and Chen, Yanfei and Pfister, Tomas and Zhang, Rui and Arik, Sercan O.},
  booktitle = {Advances in Neural Information Processing Systems (NeurIPS)},
  year      = {2024}
}

@article{wu2025resum,
  title   = {{ReSum}: Unlocking Long-Horizon Search Intelligence via Context Summarization},
  author  = {Wu, Xixi and Li, Kuan and Zhao, Yida and Jiang, Yong and Xie, Pengjun and Huang, Fei and Zhou, Jingren and others},
  journal = {arXiv preprint arXiv:2509.13313},
  year    = {2025}
}

@article{sun2025contextfolding,
  title   = {Scaling Long-Horizon {LLM} Agent via Context-Folding},
  author  = {Sun, Weiwei and Lu, Miao and Ling, Zhan and others},
  journal = {arXiv preprint arXiv:2510.11967},
  year    = {2025}
}

@article{ye2025agentfold,
  title   = {{AgentFold}: Long-Horizon Web Agents with Proactive Context Management},
  author  = {Ye, Rui and Zhang, Zhongwang and Li, Kuan and others},
  journal = {arXiv preprint arXiv:2510.24699},
  year    = {2025}
}

@article{kang2025acon,
  title   = {{ACON}: Optimizing Context Compression for Long-Horizon {LLM} Agents},
  author  = {Kang, Minki and Chen, Wei-Ning and Han, Dongge and others},
  journal = {arXiv preprint arXiv:2510.00615},
  year    = {2025}
}

@inproceedings{shinn2023reflexion,
  title     = {Reflexion: Language Agents with Verbal Reinforcement Learning},
  author    = {Shinn, Noah and Cassano, Federico and Berman, Edward and Gopinath, Ashwin and Narasimhan, Karthik and Yao, Shunyu},
  booktitle = {Advances in Neural Information Processing Systems (NeurIPS)},
  year      = {2023}
}

@inproceedings{park2023generative,
  title     = {Generative Agents: Interactive Simulacra of Human Behavior},
  author    = {Park, Joon Sung and O'Brien, Joseph C. and Cai, Carrie J. and Morris, Meredith Ringel and Liang, Percy and Bernstein, Michael S.},
  booktitle = {ACM Symposium on User Interface Software and Technology (UIST)},
  year      = {2023}
}

@article{wang2023voyager,
  title   = {Voyager: An Open-Ended Embodied Agent with Large Language Models},
  author  = {Wang, Guanzhi and Xie, Yuqi and Jiang, Yunfan and Mandlekar, Ajay and Xiao, Chaowei and Zhu, Yuke and Fan, Linxi and Anandkumar, Anima},
  journal = {arXiv preprint arXiv:2305.16291},
  year    = {2023}
}

@inproceedings{zhao2024expel,
  title     = {{ExpeL}: {LLM} Agents Are Experiential Learners},
  author    = {Zhao, Andrew and Huang, Daniel and Xu, Quentin and Lin, Matthieu and Liu, Yong-Jin and Huang, Gao},
  booktitle = {AAAI Conference on Artificial Intelligence},
  year      = {2024}
}

@inproceedings{zhong2024memorybank,
  title     = {{MemoryBank}: Enhancing Large Language Models with Long-Term Memory},
  author    = {Zhong, Wanjun and Guo, Lianghong and Gao, Qiqi and Ye, He and Wang, Yanlin},
  booktitle = {AAAI Conference on Artificial Intelligence},
  year      = {2024}
}

@article{chhikara2025mem0,
  title   = {{Mem0}: Building Production-Ready {AI} Agents with Scalable Long-Term Memory},
  author  = {Chhikara, Prateek and Khant, Dev and Aryan, Saket and Singh, Taranjeet and Yadav, Deshraj},
  journal = {arXiv preprint arXiv:2504.19413},
  year    = {2025}
}

@inproceedings{wang2024memoryllm,
  title     = {{MemoryLLM}: Towards Self-Updatable Large Language Models},
  author    = {Wang, Yu and Gao, Yifan and Chen, Xiusi and Jiang, Haoming and Li, Shiyang and Yang, Jingfeng and Yin, Qingyu and Li, Zheng and Li, Xian and Yin, Bing and Shang, Jingbo and McAuley, Julian},
  booktitle = {International Conference on Machine Learning (ICML)},
  year      = {2024}
}

@article{yu2025memagent,
  title   = {{MemAgent}: Reshaping Long-Context {LLM} with Multi-Conv {RL}-based Memory Agent},
  author  = {Yu, Hongli and Chen, Tinghong and Feng, Jiangtao and others},
  journal = {arXiv preprint arXiv:2507.02259},
  year    = {2025}
}

@article{wang2025memalpha,
  title   = {{Mem-$\alpha$}: Learning Memory Construction via Reinforcement Learning},
  author  = {Wang, Yu and Takanobu, Ryuichi and Liang, Zhiqi and others},
  journal = {arXiv preprint arXiv:2509.25911},
  year    = {2025}
}

@article{zhou2025mem1,
  title   = {{MEM1}: Learning to Synergize Memory and Reasoning for Efficient Long-Horizon Agents},
  author  = {Zhou, Zijian and Qu, Ao and Wu, Zhaoxuan and Kim, Sunghwan and Prakash, Alok and Rus, Daniela and Zhao, Jinhua and Low, Bryan Kian Hsiang and Liang, Paul Pu},
  journal = {arXiv preprint arXiv:2506.15841},
  year    = {2025}
}

@inproceedings{weston2015memnn,
  title     = {Memory Networks},
  author    = {Weston, Jason and Chopra, Sumit and Bordes, Antoine},
  booktitle = {International Conference on Learning Representations (ICLR)},
  year      = {2015}
}

@inproceedings{lewis2020rag,
  title     = {Retrieval-Augmented Generation for Knowledge-Intensive {NLP} Tasks},
  author    = {Lewis, Patrick and Perez, Ethan and Piktus, Aleksandra and Petroni, Fabio and Karpukhin, Vladimir and Goyal, Naman and K{\"u}ttler, Heinrich and Lewis, Mike and Yih, Wen-tau and Rockt{\"a}schel, Tim and Riedel, Sebastian and Kiela, Douwe},
  booktitle = {Advances in Neural Information Processing Systems (NeurIPS)},
  year      = {2020}
}

@inproceedings{guu2020realm,
  title     = {{REALM}: Retrieval-Augmented Language Model Pre-Training},
  author    = {Guu, Kelvin and Lee, Kenton and Tung, Zora and Pasupat, Panupong and Chang, Ming-Wei},
  booktitle = {International Conference on Machine Learning (ICML)},
  year      = {2020}
}

@inproceedings{borgeaud2022retro,
  title     = {Improving Language Models by Retrieving from Trillions of Tokens},
  author    = {Borgeaud, Sebastian and Mensch, Arthur and Hoffmann, Jordan and Cai, Trevor and Rutherford, Eliza and Millican, Katie and others},
  booktitle = {International Conference on Machine Learning (ICML)},
  year      = {2022}
}

@article{izacard2023atlas,
  title   = {Atlas: Few-shot Learning with Retrieval Augmented Language Models},
  author  = {Izacard, Gautier and Lewis, Patrick and Lomeli, Maria and Hosseini, Lucas and Petroni, Fabio and Schick, Timo and Dwivedi-Yu, Jane and Joulin, Armand and Riedel, Sebastian and Grave, Edouard},
  journal = {Journal of Machine Learning Research (JMLR)},
  year    = {2023}
}

@inproceedings{wang2024scm,
  title     = {{SCM}: Enhancing Large Language Model with Self-Controlled Memory Framework},
  author    = {Wang, Bing and Liang, Xinnian and Yang, Jian and Huang, Hui and Wu, Shuangzhi and Wu, Peihao and Lu, Lu and Ma, Zejun and Li, Zhoujun},
  booktitle = {International Conference on Database Systems for Advanced Applications (DASFAA)},
  year      = {2025}
}

@inproceedings{asai2024selfrag,
  title     = {{Self-RAG}: Learning to Retrieve, Generate, and Critique through Self-Reflection},
  author    = {Asai, Akari and Wu, Zeqiu and Wang, Yizhong and Sil, Avirup and Hajishirzi, Hannaneh},
  booktitle = {International Conference on Learning Representations (ICLR)},
  year      = {2024}
}

@inproceedings{fountas2025emllm,
  title     = {Human-inspired Episodic Memory for Infinite Context {LLMs}},
  author    = {Fountas, Zafeirios and Benfeghoul, Martin A. and Oomerjee, Adnan and Christopoulou, Fenia and Lampouras, Gerasimos and Bou-Ammar, Haitham and Wang, Jun},
  booktitle = {International Conference on Learning Representations (ICLR)},
  year      = {2025}
}

@inproceedings{gutierrez2024hipporag,
  title     = {{HippoRAG}: Neurobiologically Inspired Long-Term Memory for Large Language Models},
  author    = {Guti{\'e}rrez, Bernal Jim{\'e}nez and Shu, Yiheng and Gu, Yu and Yasunaga, Michihiro and Su, Yu},
  booktitle = {Advances in Neural Information Processing Systems (NeurIPS)},
  year      = {2024}
}

@inproceedings{gutierrez2025hipporag2,
  title     = {From {RAG} to Memory: Non-Parametric Continual Learning for Large Language Models},
  author    = {Guti{\'e}rrez, Bernal Jim{\'e}nez and Shu, Yiheng and Qi, Weijian and Zhou, Sizhe and Su, Yu},
  booktitle = {International Conference on Machine Learning (ICML)},
  year      = {2025}
}

@article{edge2024graphrag,
  title   = {From Local to Global: A Graph {RAG} Approach to Query-Focused Summarization},
  author  = {Edge, Darren and Trinh, Ha and Cheng, Newman and Bradley, Joshua and Chao, Alex and Mody, Apurva and Truitt, Steven and Larson, Jonathan},
  journal = {arXiv preprint arXiv:2404.16130},
  year    = {2024}
}

@article{guo2024lightrag,
  title   = {{LightRAG}: Simple and Fast Retrieval-Augmented Generation},
  author  = {Guo, Zirui and Xia, Lianghao and Yu, Yanhua and Ao, Tu and Huang, Chao},
  journal = {arXiv preprint arXiv:2410.05779},
  year    = {2024}
}

@article{rasmussen2025zep,
  title   = {{Zep}: A Temporal Knowledge Graph Architecture for Agent Memory},
  author  = {Rasmussen, Preston and Paliychuk, Pavlo and Beauvais, Travis and Ryan, Jack and Chalef, Daniel},
  journal = {arXiv preprint arXiv:2501.13956},
  year    = {2025}
}

@article{chen2026knowu,
  title={KnowU-Bench: Towards Interactive, Proactive, and Personalized Mobile Agent Evaluation},
  author={Chen, Tongbo and Lu, Zhengxi and Xu, Zhan and Shao, Guocheng and Zhao, Shaohan and Tang, Fei and Du, Yong and Song, Kaitao and Liu, Yizhou and Yan, Yuchen and others},
  journal={arXiv preprint arXiv:2604.08455},
  year={2026}
}

@article{deepseekv4,
  author = {{DeepSeekAI}},
  title  = {DeepSeek-V4 Model Card},
  year   = {2026},
  url    = {huggingface.co/deepseek-ai/DeepSeek-V4-Pro}
}

@article{claudeopus46,
  author = {{Anthropic}},
  title  = {Claude Opus 4.6 System Card},
  year   = {2026},
  url    = {https://www.anthropic.com/claude-opus-4-6-system-card}
}

@article{glm51,
  author = {{Z.ai}},
  title  = {{GLM-5.1} Model Card},
  year   = {2026},
  url    = {https://huggingface.co/zai-org/GLM-5.1}
}

@article{doubao2025seed20,
  author = {{ByteDance Seed}},
  title  = {Seed 2.0 Model Card: Towards Intelligence Frontier for Real-World Complexity},
  year   = {2026},
  url    = {seed.bytedance.com/en/seed2}
}

@article{kimi26,
  author = {{Moonshot AI}},
  title  = {{Kimi-K2.6} Model Card},
  year   = {2026},
  url    = {https://huggingface.co/moonshotai/Kimi-K2.6}
}

@article{minimax27,
  author = {{MiniMax}},
  title  = {{MiniMax-M2.7}: Model Self-Improvement, Driving Productivity Innovation Through Technological Breakthroughs},
  year   = {2026},
  url    = {https://www.minimax.io/models/text/m27}
}

@article{wang2026agentnoisebench,
  title={Agentnoisebench: Benchmarking robustness of tool-using llm agents under noisy condition},
  author={Wang, Ruipeng and Chen, Yuxin and Wang, Yukai and Wu, Chang and Fang, Junfeng and Cai, Xiaodong and Gu, Qi and Su, Hui and Zhang, An and Wang, Xiang and others},
  journal={arXiv preprint arXiv:2602.11348},
  year={2026}
}

@article{zheng2026risky,
  title={Risky-Bench: Probing Agentic Safety Risks under Real-World Deployment},
  author={Zheng, Jingnan and Luo, Yanzhen and Xu, Jingjun and Liu, Bingnan and Chen, Yuxin and Cui, Chenhang and Deng, Gelei and Lu, Chaochao and Wang, Xiang and Zhang, An and others},
  journal={arXiv preprint arXiv:2602.03100},
  year={2026}
}

@article{shi2026aj,
  title={AJ-Bench: Benchmarking Agent-as-a-Judge for Environment-Aware Evaluation},
  author={Shi, Wentao and Wang, Yu and Zhao, Yuyang and Chen, Yuxin and Feng, Fuli and Hao, Xueyuan and Su, Xi and Gu, Qi and Su, Hui and Cai, Xunliang and others},
  journal={arXiv preprint arXiv:2604.18240},
  year={2026}
}

\clearpage
\onecolumn

\renewcommand{\cftsecfont}{\normalsize} 
\renewcommand{\cftsubsecfont}{\normalsize} 
\renewcommand{\cftbeforesecskip}{12pt}      
\renewcommand{\cftbeforesubsecskip}{12pt}  
\renewcommand{\cftbeforesubsubsecskip}{6pt}

\renewcommand{\contentsname}{Appendix}
\addtocontents{toc}{\protect\setcounter{tocdepth}{3}}
\appendix
{\hypersetup{linkcolor=RoyalBlue}
\tableofcontents
}
\clearpage

\appendix
\clearpage

\section{Benchmark Construction}
\label{app:benchmark_data}

VitaBench 2.0 is designed to model long-term, user-centric interaction scenarios, where agents are required to continuously satisfy long-term user needs. The benchmark is built around 56 curated users, each associated with a temporally ordered sequence of tasks spanning diverse real-world domains. This design enables systematic evaluation of preference inference, preference evolution, and proactive decision-making in realistic settings.
In the following, we describe the construction of user profiles, preferences, interaction histories, and task environments in detail.

\subsection{User}
\label{app:user}

Modeling realistic users is critical in our setting, as agents are required to infer and adapt to evolving user preferences over time. However, due to well-known biases and hallucination issues, large language models alone are insufficient for generating high-quality user data. To ensure realism and consistency, we rely on manual annotation for user profiles and preferences, complemented by controlled synthesis for interaction histories.

\subsubsection{User Profile}
Each user is associated with a detailed profile describing their demographic attributes and background information. These profiles are manually curated and inspired by our real-world application scenarios to ensure diversity and realism. The profiles serve as the foundation for preference construction and downstream task generation.
We provide an illustrative example of user profile below.

\begin{tcolorbox}[
  colback=gray!5!white, colframe=gray!75!black, breakable,
  title=Example User Profile]
\small
\texttt{user\_id}: T850685 \\
\texttt{occupation}: Teacher \\
\texttt{gender}: Female \\
\texttt{date\_of\_birth}: 1991-08-16 \\
\texttt{current\_city}: Leshan, Sichuan Province \\
\texttt{home\_address}: Building 1, Unit 602, Nvshen Yuan, Shizhong District, Leshan, Sichuan Province \\
\texttt{work\_address}: Leshan Normal University, Leshan, Sichuan Province \\
\texttt{hometown}: Yibin, Sichuan Province \\
\texttt{education}: Doctoral degree \\
\texttt{family\_status}: [``Unmarried, no children''] \\
\texttt{chronic\_conditions\_allergies}: [``Chronic rhinitis'']
\end{tcolorbox}

\subsubsection{User Preference}
\label{app:user_preference}
Each user is associated with a set of preferences expressed as natural language statements (e.g., dietary restrictions, spending habits, travel styles), grounded in the corresponding user profile. To ensure realism and consistency, all user preferences are manually annotated.
Preferences cover diverse aspects of daily life and vary significantly across users, resulting in a rich and heterogeneous preference space. In total, we curate over 1,000 user-specific preferences, with each user exhibiting multiple fine-grained constraints that jointly influence decision-making.
User preferences in real life are inherently dynamic. To simulate realistic evolution over time, we introduce temporally grounded changes, including preference addition (emergence of new preferences), modification (shifts in existing preferences), and deletion (disappearance of previously relevant preferences). These changes are distributed across the task sequence to reflect long-term user dynamics.
In addition, we explicitly distinguish \emph{conditional preferences}, where the correct decision depends on context that is not directly observable from the current query or maintained user preference (e.g., time, companion, or situational constraints). Such preferences require agents to recognize ambiguity and actively acquire missing information from the user, forming the basis for proactive tasks in our benchmark.
We provide an illustrative example of user preference below.

\begin{tcolorbox}[
  colback=gray!5!white, colframe=gray!75!black, breakable,
  title=Example User Preference]
\small
\texttt{user\_id}: T850685 \\
\\
\texttt{dietary\_preferences}: \\
\quad ``Likes Sichuan cuisine'' \\
\quad ``Likes sushi'' \\
\quad ``Dislikes Cantonese cuisine'' \\
\quad ``Likes fruit tea'' \\
\quad ``Prefers drinks half-sugar'' \\
\quad ``Prefers drinks ice-free'' \\
\quad ``Likes to try newly opened restaurants on weekends'' \\
\\
\texttt{shopping\_preferences}: \\
\quad ``Likes clothing with pocket designs'' \\
\quad ``Unwilling to pay a premium for luxury brands'' \\
\quad ``Likes flat shoes'' \\
\quad ``Likes to buy practical items'' \\
\quad ``Likes cool-toned colors'' \\
\\
\texttt{travel\_and\_accommodation\_preferences}: \\
\quad ``Prefers high-speed rail for travel'' \\
\quad ``Prefers chain hotel brands'' \\
\quad ``Prefers sea-view hotel rooms'' \\
\quad ``Prefers hotel rooms with windows'' \\
\\
\texttt{leisure\_preferences}: \\
\quad ``Likes playing mahjong'' \\
\quad ``Likes reading'' \\
\\
\texttt{preference\_tag\_change\_history} (selected entries across the task sequence): \\
\quad \{``content'': ``Prefers watermelon-flavored fruit tea'', ``type'': ``update'', ``source'': ``Likes fruit tea''\} \\
\quad \{``content'': ``Prefers fresh-milk--based desserts'', ``type'': ``update'', ``source'': ``Prefers local brands for desserts and snacks''\} \\
\quad \{``content'': ``Likes sushi'', ``type'': ``delete'', ``source'': ``''\} \\
\quad \{``content'': ``Likes to buy practical items'', ``type'': ``delete'', ``source'': ``''\} \\
\quad \{``content'': ``Prefers small private rooms in mahjong parlors'', ``type'': ``update'', ``source'': ``Prefers large private rooms in mahjong parlors''\} \\
\quad \{``content'': ``Wants to travel abroad'', ``type'': ``update'', ``source'': ``Does not like overseas trips''\} \\
\quad \{``content'': ``Dislikes Cantonese cuisine'', ``type'': ``delete'', ``source'': ``''\} \\
\quad \{``content'': ``Likes clothing with pocket designs'', ``type'': ``delete'', ``source'': ``''\} \\
\quad \{``content'': ``Prefers high-speed rail for travel'', ``type'': ``delete'', ``source'': ``''\} \\
\quad \{``content'': ``Likes playing mahjong'', ``type'': ``delete'', ``source'': ``''\} \\
\quad \{``content'': ``Prefers executive-floor hotel rooms'', ``type'': ``update'', ``source'': ``Prefers sea-view hotel rooms''\}
\end{tcolorbox}

\begin{tcolorbox}[
    colback=gray!5!white, colframe=gray!75!black, breakable,
    title=Example Proactive Preference]
  \small
  \texttt{user\_id}: B865629 \\
  \\
  \texttt{conditional\_preference}: \\
  \quad ``Prefers high-caffeine coffee in the morning; low-caffeine in the afternoon'' \\
  \\
  \texttt{historical\_conversation} (source of the conditional preference): \\
  \quad ``I find I'm quite sensitive to caffeine --- high-caffeine in the morning is totally fine, gives a real kick, but in the afternoon I can
  only handle low-caffeine.'' \\
  \\
  \texttt{user\_query}: \\
  \quad ``I have a meeting on the 26th, order me a coffee in advance to stay alert.'' \\
  \\
  \texttt{unresolved\_ambiguity}: \\
  \quad The correct caffeine level depends on the \emph{time} of the meeting, which is not specified in the query. The agent should proactively
  ask the user when the meeting is scheduled. \\
  \\
  \texttt{user\_clarification}: \\
  \quad ``The meeting is in the afternoon.'' \\
  \\
  \texttt{expected\_outcome}: \\
  \quad ``The ordered product should be coffee'' \\
  \quad ``The ordered product should be low-caffeine'' \\
  \quad ``Delivery address: user's workplace''
  \end{tcolorbox}

\subsubsection{Interaction History}
User preferences are not directly exposed to the agent, but are instead implicitly encoded in fragmented interaction histories. Inspired by information accessibility in real-world scenarios, we construct interaction histories consisting of two modalities: (1)~\emph{dialogues}, including multi-turn conversational interactions between users and agents, and (2)~\emph{behaviors}, including user logs such as browsing, ordering, reviewing, and searching.

To ensure both realism and sufficient difficulty, interaction histories are constructed through a controlled synthesis process guided by manually designed preference signals. Specifically, given initial user preferences, we first manually design how these preferences can be embedded into fragmented interactions and then generate a large set of interactions that implicitly reflect these preferences. As tasks are temporally ordered, additional interaction histories are generated between consecutive tasks, capturing preference evolution over time. These histories may encode preference addition, preference drift, or preference disappearance. 
We distinguish between two types of preference changes. In some cases, preference changes are explicitly reflected through interactions (e.g., a user repeatedly ordering vegetarian meals). In other cases, preference changes are not directly observable from interactions and are instead modeled as implicit state transitions. For example, a user may temporarily prefer lighter meals due to illness, or stop exhibiting pregnancy-related preferences after the corresponding period ends. 
Such implicit changes introduce additional difficulty, as agents cannot rely solely on observable interaction signals and must maintain a consistent and adaptive representation of user preferences over time.
To further increase difficulty and realism, not all interactions are preference-relevant. We deliberately introduce noise, including irrelevant actions, ambiguous signals, and short-term preference fluctuations that may appear inconsistent. This requires agents to distinguish stable preferences from noisy observations. 
All interaction histories are manually reviewed and refined to ensure that they are coherent, free of contradictions, and do not introduce unintended preference leakage. We provide an illustrative example of interaction history below.

\begin{tcolorbox}[
  colback=gray!5!white, colframe=gray!75!black, breakable,
  title=Example User Interaction History]
\small
\textbf{User ID:} \texttt{T850685} \\
\textbf{Period:} 2026-04-14 to 2026-04-15 \\
\textbf{Preference State Before:} likes fruit tea; prefers drinks half-sugar and no ice. \\
{\color{purple}\textbf{Preference Drift:} update from ``likes fruit tea'' to ``prefers watermelon-flavored fruit tea'', triggered on 2026-04-14 by the user's statement: ``Fruit teas feel sweeter these days; only watermelon still tastes refreshing. From now on, prioritize watermelon flavor for fruit tea.''} \\
\\
\textbf{Timeline} \\
\\
\textbf{2026-04-14} \\
{\color{green!50!black}\textbf{Behaviors}} \\
\quad Order: Mango pomelo sago, half-sugar and no ice, from A Little Tea (Wanda Store), 18 RMB. \\
\\
{\color{blue}\textbf{Dialogue}} \\
\quad {\color{red}User: What model are you?} \\
\quad {\color{red}Assistant: I am your AI life assistant. What can I help you with?} \\
\quad User: Just asking. Help me order mango pomelo sago from A Little Tea, the usual way. \\
\quad Assistant: Sure. Mango pomelo sago from A Little Tea Wanda Store, 18 RMB, half-sugar and no ice. Should I place the order? \\
\quad User: Go ahead. \\
\quad Assistant: Ordered. It should arrive in about 30 minutes. \\
\quad {\color{purple}User: I feel like fruit teas are getting sweeter these days; only watermelon still tastes refreshing.} \\
\quad {\color{purple}Assistant: Mango and coconut-based drinks can indeed taste rich even at half sugar, while watermelon tea is usually lighter. Should I prioritize watermelon flavors for fruit tea from now on?} \\
\quad {\color{purple}User: Yes, prioritize watermelon flavor for fruit tea in the future.} \\
\quad {\color{purple}Assistant: Got it. I will prioritize watermelon-flavored fruit tea for you from now on.} \\
\\
\textbf{2026-04-15} \\
{\color{green!50!black}\textbf{Behaviors}} \\
\quad {\color{red}Search: ``is pearl cream actually made from pearls?''} \\
\quad {\color{purple}Order: Watermelon boba fruit tea, half-sugar and no ice, from Shuyi Tealicious (Teachers College Store), 14 RMB.} \\
\\
{\color{blue}\textbf{Dialogue}} \\
\quad {\color{red}User: Is pearl cream really made from pearls?} \\
\quad {\color{red}Assistant: Traditional pearl cream often contains pearl powder, but the amount varies a lot across products today.} \\
\quad {\color{red}User: Oh, I was just curious.} \\
\quad {\color{purple}User: By the way, help me order a fruit tea from Shuyi.} \\
\quad {\color{purple}Assistant: Shuyi Tealicious Teachers College Store has ``Watermelon Boba Fruit Tea'' for 14 RMB, half-sugar and no ice, delivered to Leshan Teachers College. Does that work?} \\
\quad {\color{purple}User: Yes, place the order.} \\
\quad {\color{purple}Assistant: Ordered. It should arrive in about 20 minutes.} \\
\\
\textbf{2026-04-17} \\
\textbf{......}
\end{tcolorbox}

\subsubsection{Prompt Template}
We provide the prompt templates used for synthesizing user interaction histories below.

\begin{tcolorbox}[
  colback=gray!5!white, colframe=gray!75!black, breakable,
  title=User Interaction Synthesis Prompt Template]
\small
\# General Prompt \\
I will provide you with information about a real human user and a period of time during which the user's personalized preferences have changed. The user fulfills everyday needs on a comprehensive life-service platform, either by interacting with an AI agent or by taking platform actions independently. Based on the input fields, synthesize a temporally ordered interaction history from \texttt{start\_date} to \texttt{end\_date}. The history should contain {\color{blue}user-agent dialogues}, {\color{green!50!black}platform behavior logs}, {\color{purple}preference drift signals}, and {\color{red}preference-irrelevant noise}. The central goal is to make the preference changes naturally observable through the generated dialogues and behaviors, while keeping all personalized content grounded in the provided user profile and current preferences. \\
\\
\# Input Fields \\
1. \texttt{start\_date}: the first date covered by the generated history. \\
2. \texttt{end\_date}: the last date covered by the generated history. \\
3. \texttt{user\_profile}: the user's demographic background, lifestyle context, and other basic information. \\
4. \texttt{current\_preferences}: the user's complete preference state after the drift events in this period. \\
5. \texttt{preference\_changes}: the preference additions, deletions, or modifications that occur during this period. These changes are the core signals that must be reflected. \\
6. \texttt{historical\_chat\_list}: user utterance fragments that must all appear in the generated dialogues. They may be placed on different dates and adapted to fit the context, but their core meaning must be preserved. \\
7. \texttt{historical\_behavior\_text}: behavior records that must all appear in the generated platform behavior logs. They may be distributed across different dates. \\
8. \texttt{noise\_text}: preference-irrelevant content that must be naturally incorporated into the generated dialogues or behaviors. \\
\\
{\color{purple}
\# Preference Drift \\
Preference drift is the primary target of this history. The generated dialogues and behaviors should naturally reflect how the user's preferences change during this period. Newly added preferences should appear through attempts, discoveries, explicit interest, or repeated related actions. Modified preferences should show a plausible transition from the previous tendency to the new one. Deleted preferences should be reflected through reduced interest, explicit rejection, or behavior that indicates the preference is no longer active. These drift signals should be distributed across the history rather than stated mechanically in one place. \\
}
\\
{\color{blue}
\# Dialogues \\
Dialogues represent multi-turn interactions between the user and the AI agent. Each topic or intent should be sufficiently developed: task-oriented dialogues such as ordering food, booking tickets, shopping, or travel planning should usually contain 4--8 user-assistant turns, while casual conversations should usually contain 3--6 turns. The user's language style should match the user profile and preference descriptions, and all required utterance fragments in \texttt{historical\_chat\_list} must be included as user messages. The agent's responses should be varied and natural, and a dialogue should not end before the user's intent has been fully expressed or resolved. \\
}
\\
{\color{green!50!black}
\# Behaviors \\
Behavior records represent actions taken by the user on the platform, including actions completed independently and actions related to prior agent interactions. All required behaviors in \texttt{historical\_behavior\_text} must be included. Not every behavior needs a corresponding dialogue; independent searches, browsing, favorites, comments, ratings, and orders may appear as standalone platform activity. Supported behavior types include search, order, rate, comment, high\_freq\_browse, complaint, add\_to\_cart, and favorite. The order type covers delivery, ticket, instore, travel\_ticket, hotel, attraction, local\_trip, and errand scenarios. \\
}
\\
{\color{red}
\# Noise \\
Noise refers to content that is unrelated to personalization but makes the history more realistic. All items in \texttt{noise\_text} must be naturally incorporated into the generated dialogues or behavior logs. Noise may include casual chat, ordinary one-off actions, ambiguous records, or preference-irrelevant interactions. It should not introduce new user preferences, contradict the user profile, or distract from the preference drift signals. \\
}
\\
\# Notes \\
All records must occur between \texttt{start\_date} and \texttt{end\_date}, be organized by day, and appear in increasing chronological order. Each day may contain any number of dialogues or behaviors, including days with only one type of record. \\
All personalized content must be traceable to \texttt{current\_preferences}, \texttt{preference\_changes}, or \texttt{user\_profile}. Do not invent preferences, habits, interests, or consumption patterns that were not provided. You may add extra dialogues and behaviors to reflect unchanged preferences and make the history richer, but they must remain grounded. Merchant, restaurant, and hotel names may be plausibly fabricated, but real-world attractions, cuisines, aircraft models, and well-known brands must not be fabricated. \\
Return only a JSON array. The response must start with \texttt{[} and end with \texttt{]}. Do not wrap the output in a code block. If quotation marks are needed inside a JSON string value, use Chinese quotation marks such as \texttt{[]}, not unescaped English double quotation marks. \\
\\
\# Output Format \\
Return a list of daily records. Each record contains a date, a behavior list, and a dialogue list. Use the following structure:
\begin{verbatim}
[{"date":"2023-07-07",
  "behavior":[
    {"behavior_type":"search",
     "content":{"keyword":"swimming pool"}},
    {"behavior_type":"order",
     "content":{"scenario":"instore",
                "merchant_name":"a fitness center",
                "tags":["heated pool"],
                "items":[{"product_name":"annual pass",
                          "price":5888,
                          "quantity":1}]}}],
  "dialogue":[
    {"role":"user","content":"Where is a nearby swimming pool?"},
    {"role":"assistant","content":"I found three options ..."},
    {"role":"user","content":"I prefer a heated pool."},
    {"role":"assistant","content":"The first option has one ..."}]},
 {"date":"2023-07-15",
  "behavior":[
    {"behavior_type":"rate",
     "content":{"target_name":"a fitness center",
                "target_type":"merchant",
                "score":4.5}}],
  "dialogue":[]}]
\end{verbatim}
\# Inputs
\begin{verbatim}
{"start_date": "{start_date}",
 "end_date": "{end_date}",
 "user_profile": {user_profile},
 "current_preferences": {current_preferences},
 "preference_changes": {preference_changes},
 "historical_chat_list": {historical_chat_list},
 "historical_behavior_text": {historical_behavior_text},
 "noise_text": {noise_text}}
\end{verbatim}
\end{tcolorbox}

\subsection{Task Environment}
\label{app:environment}

The environment in VitaBench 2.0 provides the execution space in which the agent interacts with tools to fulfill user requests. As each task is formulated as an agentic task, the environment must expose a structured and realistic candidate space that supports tool-based decision making. 
We construct domain-specific environments that simulate real-world service scenarios, together with executable tool interfaces and structured data. 
\subsubsection{Toolset}
We adopt the toolset design from VitaBench~\cite{he2025vitabench}, covering three representative domains: Delivery, In-store Consumption, and Online Travel Agency. In total, the benchmark includes 66 tools that expose structured APIs for retrieving and manipulating environment states.
The tools are designed to be consistent across domains and sufficiently expressive to support multi-step interactions, while avoiding unnecessary complexity. This ensures that agents must correctly invoke tools to access relevant information, but that task difficulty primarily stems from preference inference and decision-making rather than tool usage itself.
We provide an illustrative example of toolset below.
\begin{tcolorbox}[
  colback=gray!5!white, colframe=gray!75!black, breakable,
  title=Example Toolset]
\small
\textbf{Delivery (12 tools).} \\
\texttt{delivery\_distance\_to\_time}, \texttt{get\_delivery\_store\_info}, \texttt{get\_delivery\_product\_info}, \texttt{delivery\_store\_search\_recommand}, \texttt{delivery\_product\_search\_recommand}, \texttt{create\_delivery\_order}, \texttt{pay\_delivery\_order}, \texttt{get\_delivery\_order\_status}, \texttt{cancel\_delivery\_order}, \texttt{modify\_delivery\_order}, \texttt{search\_delivery\_orders}, \texttt{get\_delivery\_order\_detail}. \\
\\
\textbf{In-store (16 tools).} \\
\texttt{instore\_shop\_search\_recommend}, \texttt{instore\_product\_search\_recommend}, \texttt{create\_instore\_product\_order}, \texttt{pay\_instore\_order}, \texttt{instore\_cancel\_order}, \texttt{instore\_book}, \texttt{pay\_instore\_book}, \texttt{instore\_cancel\_book}, \texttt{instore\_reservation}, \texttt{instore\_modify\_reservation}, \texttt{instore\_cancel\_reservation}, \texttt{get\_instore\_orders}, \texttt{get\_instore\_reservations}, \texttt{get\_instore\_books}, \texttt{search\_instore\_book}, \texttt{search\_instore\_reservation}. \\
\\
\textbf{OTA (37 tools).} \\
\texttt{get\_ota\_hotel\_info}, \texttt{get\_ota\_attraction\_info}, \texttt{get\_ota\_flight\_info}, \texttt{get\_ota\_train\_info}, \texttt{hotel\_search\_recommand}, \texttt{attractions\_search\_recommend}, \texttt{flight\_search\_recommend}, \texttt{train\_ticket\_search}, \texttt{create\_hotel\_order}, \texttt{create\_attraction\_order}, \texttt{create\_flight\_order}, \texttt{create\_train\_order}, \texttt{pay\_hotel\_order}, \texttt{pay\_attraction\_order}, \texttt{pay\_flight\_order}, \texttt{pay\_train\_order}, \texttt{search\_hotel\_order}, \texttt{search\_attraction\_order}, \texttt{search\_flight\_order}, \texttt{search\_train\_order}, \texttt{get\_hotel\_order\_detail}, \texttt{get\_attraction\_order\_detail}, \texttt{get\_flight\_order\_detail}, \texttt{get\_train\_order\_detail}, \texttt{modify\_train\_order}, \texttt{modify\_flight\_order}, \texttt{cancel\_hotel\_order}, \texttt{cancel\_attraction\_order}, \texttt{cancel\_flight\_order}, \texttt{cancel\_train\_order}, \texttt{get\_taxi\_info}, \texttt{taxi\_search}, \texttt{create\_taxi\_order}, \texttt{pay\_taxi\_order}, \texttt{cancel\_taxi\_order}, \texttt{search\_taxi\_order}, \texttt{get\_taxi\_order\_detail}. \\
\\
\textbf{Personalization (2 tools).} \\
\texttt{read\_preference\_memory}, \texttt{query\_preference\_memory}. \\
\\
\textbf{Illustrative tool schemas} (taken verbatim from the codebase; we keep the original spelling \texttt{recommand} for consistency with the data files). \\
\\
\texttt{delivery\_store\_search\_recommand}: \\
\quad \texttt{description}: ``In delivery scenario, can extract keywords describing stores from user expressions, search or recommend multiple stores'' \\
\quad \texttt{preconditions}: ``In delivery scenario, get keywords describing stores'' \\
\quad \texttt{postconditions}: ``Return store list, guide user to select and confirm store'' \\
\quad \texttt{args}: \{``keywords'': ``Keywords describing stores''\} \\
\quad \texttt{returns}: ``Structured store information output'' \\
\quad \texttt{tool\_type}: ``READ'' \\
\\
\texttt{create\_delivery\_order}: \\
\quad \texttt{description}: ``Create delivery order, only supports single store ordering, single store can order multiple items'' \\
\quad \texttt{preconditions}: ``In delivery scenario, determine unique store id and one or more product ids, determine user's dietary restrictions, reflect in the order'' \\
\quad \texttt{postconditions}: ``Return order information, ask user whether to pay the order'' \\
\quad \texttt{args}: \{``user\_id'', ``store\_id'', ``food\_ids'', ``food\_cnts'', ``address'', ``dispatch\_time'', ``attributes'', ``note''\} \\
\quad \texttt{returns}: ``Order information'' \\
\quad \texttt{tool\_type}: ``WRITE'' \\
\\
\texttt{query\_preference\_memory}: \\
\quad \texttt{description}: ``Query user preference memory with a specific question'' \\
\quad \texttt{preconditions}: ``Use when you need to query specific user preference information'' \\
\quad \texttt{postconditions}: ``Returns preference information relevant to the query'' \\
\quad \texttt{args}: \{``query'': ``Natural language question about user preferences''\} \\
\quad \texttt{returns}: ``Preference information relevant to the query'' \\
\quad \texttt{tool\_type}: ``READ''
\end{tcolorbox}

\subsubsection{Environment Synthesis}

As each task in VitaBench~2.0 is formulated as an agentic task, it requires an executable environment in which the agent can invoke tools to interact and fulfill user needs (e.g., placing a food-delivery order). Our benchmark contains thousands of tasks spanning diverse domains (e.g., food delivery, in-store services, and travel booking), and ensuring sufficient task difficulty requires each environment to present a rich and structured candidate space. Manually constructing such environments at scale is therefore impractical. To address this, we design a synthesis pipeline that generates structured, executable environments for each task, complemented by programmatic validation and manual refinement to ensure correctness and consistency.

We adopt a multi-agent design that decomposes environment synthesis into a sequence of specialized components, each responsible for a well-defined subtask. Given a user profile, user query, and evaluation rubric, the environment generator materializes a database of merchants and items consistent with the task specification, including both valid candidates and carefully constructed distractors. Importantly, our benchmark evaluates not only basic tool-use capabilities but also the ability to infer and leverage user preferences. Accordingly, we construct environments whose difficulty arises from two complementary dimensions.
First, conditioned on the user query, we generate merchants and items that are relevant to the domain but do not satisfy the task constraints, requiring the agent to correctly invoke tools and filter out irrelevant candidates. Second, conditioned on the evaluation rubric, we generate candidates that satisfy the user query but violate user preferences, requiring the agent to infer and apply preference information to eliminate such distractors. 

To improve realism and control complexity, environments are synthesized in a top-down manner: we first generate a set of merchants and then populate each merchant with items. We explicitly control the number of items that satisfy all rubric constraints to ensure that each task admits a well-defined solution while remaining sufficiently challenging. To ensure evaluation correctness, we first apply a strong model to verify the logical consistency of the environment and detect potential conflicts, followed by human expert review to further refine the data and ensure that the resulting environments are both valid and non-trivial.
We provide an illustrative example of task environment below.

\begin{tcolorbox}[
  colback=gray!5!white, colframe=gray!75!black, breakable,
  title=Example Environment]
\small
The following example is adapted from the first subtask of user \texttt{M793481}. The user asks the agent to book a ticket from Chengdu to Kunming for the coming Saturday. The environment contains weather and location context, together with a train-candidate database. Only one weather record, one distractor train, and one target train are shown; omitted entries are denoted by \texttt{...}.

\begin{verbatim}
{
  "task_turn_num": "M793481_01",
  "domain": "ota",
  "instruction": "I want to visit Kunming. Please book a ticket
                  for this Saturday.",
  "current_time": "2024-07-02",
  "task": "OTA / ticket booking",
  "rubric": [
    "the recommended product should include a ticket for 2024-07-06",
    "the recommended product should be a bullet-train ticket",
    "the recommended product should depart from Chengdu",
    "the recommended product should arrive in Kunming"
  ],
  "environment": {
    "time": "2024-07-02 14:30:00",
    "weather": [
      {"city": "Kunming", "category": "sunny",
       "datetime": "2024-07-06", "temperature": [18, 26],
       "humidity": 58},
      "..."
    ],
    "location": [
      {"address": "Chunxi Road, Jinjiang District, Chengdu",
       "longitude": 104.0665, "latitude": 30.5728}
    ],
    "trains": {
      "S17775666762126018_T00002": {
        "train_number": "G2815",
        "departure_city": "Chongqing",
        "arrival_city": "Kunming",
        "departure_time": "2024-07-06 08:20",
        "arrival_time": "2024-07-06 13:45",
        "tags": ["high-speed rail", "fast", "comfortable"],
        "train_type": "distraction",
        "distraction_reason": "departure city is Chongqing, violating
                               the Chengdu-departure requirement",
        "products": [
          {"seat_type": "second class", "price": 398, "quantity": 68,
           "date": "2024-07-06", "product_type": "distraction",
           "product_id": "S17775666762126018_P00003"},
          {"seat_type": "first class", "price": 628, "quantity": 25,
           "date": "2024-07-06", "product_type": "distraction",
           "product_id": "S17775666762126018_P00004"}
        ]
      },
      "...": "other query-relevant distractor trains are omitted"
    }
  }
}
\end{verbatim}
\end{tcolorbox}

\subsubsection{Prompt Template}

We provide the prompt templates for the environment generator, noise injector, and verifier below.

\begin{tcolorbox}[
  colback=gray!5!white, colframe=gray!75!black, breakable,
  title=Environment Synthesis Prompt Template]
\small
\# General Prompt \\
Produce a structured database of merchants and items that is consistent with the task specification, forming an executable environment for the agent. You are an environment synthesizer for a \texttt{\{domain\}} agentic benchmark. Your job is to produce a complete JSON environment database that turns the given task into an executable testbed, where an AI agent must use tools to find the correct answer. \\
\\
\# Task Specification \\
- User profile: \texttt{\{user\_profile\}} \\
- Current time: \texttt{\{current\_time\}} \\
- Task type: \texttt{\{task\_type\}} \\
- User query: \texttt{\{instruction\}} \\
- Evaluation rubric: \texttt{\{rubric\_formatted\}} \\
\\
\# Output Structure \\
Produce a JSON database in the following abbreviated schema. The field \texttt{\{entity\_key\}} should be one of \texttt{stores}, \texttt{shops}, \texttt{trains}, \texttt{hotels}, \texttt{flights}, or \texttt{attractions}, depending on the domain.
\begin{verbatim}
{
  "time": "...",
  "weather": [...],
  "location": [...],
  "{entity_key}": {
    "entity_001": {
      "id": "...",
      "name": "...",
      "...": "domain attributes such as score, location, tags",
      "{entity_type}_type": "target | distraction",
      "distraction_reason": "...",
      "products": [
        {
          "name": "...",
          "price": ...,
          "...": "product attributes, options, tags",
          "product_type": "target | query_distractor |
                           preference_distractor | irrelevant",
          "distraction_reason": "..."
        }
      ]
    }
  }
}
\end{verbatim}
\# Top-Down Synthesis Protocol \\
Pass 1 --- Merchant layer. Generate a realistic set of merchants relevant to the user query. Merchant names, tags, scores, and locations should resemble a real search-result page in the target domain, e.g., natural names such as ``Dianwei Crossing-Bridge Rice Noodles (High-Tech Branch)'' rather than generic test names. \\
Pass 2 --- Item layer. For each merchant, populate items. The full environment must contain exactly one globally valid target product that satisfies both the user query and all rubric constraints, multiple distractors along the two axes below, and a small number of plainly irrelevant items for realism. \\
\\
\# Rubric Decomposition \\
Before generating the database, parse the evaluation rubric into atomic constraint dimensions. For example, if the rubric requires the ordered product to be crossing-bridge rice noodles and the flavor to be tomato, the dimensions are \texttt{category = crossing-bridge rice noodles} and \texttt{flavor = tomato}. Also infer the keywords that the agent is likely to search for from the user query. \\
\\
{
\# Axis A: Query-Relevant but Rubric-Violating Distractors \\
Generate items that look domain-appropriate and satisfy the user's literal request, but violate one or more rubric constraints. These items force the agent to issue the right tool calls and filter the candidate space correctly. Label them with \texttt{product\_type = "query\_distractor"}. \\
For each rubric dimension, design at least five ``only-one-thing-off'' distractors. Each such distractor should violate exactly one dimension while satisfying all other dimensions, and should be almost identical to the target in name, tags, attributes, and price range. In addition, include 3--5 multi-dimensional violations, each violating 2--3 dimensions simultaneously. \\
}
\\
{
\# Axis B: Rubric-Satisfying but Preference-Violating Distractors \\
Generate items that satisfy the user's literal query and most rubric constraints, but violate latent preferences inferred from the user profile. These items force the agent to infer and apply user preferences rather than selecting the first plausible candidate. Label them with \texttt{product\_type = "preference\_distractor"}. Distribute these distractors across multiple merchants rather than clustering them in one place. \\
}
\\
\# Distractor Annotation \\
For every non-target item, the field \texttt{distraction\_reason} must explicitly state which rubric or preference constraint is violated and how. For plainly irrelevant items, label them with \texttt{product\_type = "irrelevant"} and include a short reason. \\
\\
\# Complexity Control \\
- Target uniqueness: the target product must be globally unique. If any other item could also satisfy all rubric constraints, relabel it as a distractor. \\
- Scale: generate 10--15 merchants, 4--8 products per merchant, at least 40 products in total, and at least 30 distractors. \\
- Placement: the target product must not appear in the first five merchants; place it in the middle or later portion of the list. \\
- Ranking: the target merchant must not be the highest-scored merchant. Its score should be in [4.0, 4.6], and at least two other merchants should have higher scores. \\
- Search coverage: the names and tags of all merchants should include the main search keywords inferred from the user query, so lexical uniqueness alone is insufficient. \\
\\
\# Address and Location Constraints \\
If the rubric involves a delivery address or location, populate the \texttt{location} field with coordinates consistent with that constraint. For delivery-type tasks, all merchants should lie within a reasonable range of the specified address. \\
\\
\# Output Requirements \\
Return the JSON database only. Do not include comments, explanation text, or markdown fences outside the JSON body. Use placeholder IDs such as \texttt{store\_001} and \texttt{prod\_001}; the system will rewrite them in post-processing. The \texttt{weather} field should cover the current day $\pm$ 1--2 days.
\end{tcolorbox}

\begin{tcolorbox}[
  colback=gray!5!white, colframe=gray!75!black, breakable,
  title=Noise Injector Prompt Template]
\small
\# General Prompt \\
Given an already-generated environment, inject additional adversarial distractors so that an agent without preference memory cannot solve the task reliably, while an agent with preference memory still can. You are a noise injector for an agentic benchmark. Given a task and its initial environment, inject additional distractor entries so that an agent relying only on the literal user query cannot reliably select the correct answer, while an agent with full preference memory still can. \\
\\
\# Core Objective \\
The null agent has no user preference memory and acts purely on the literal user query. The injected distractors must appear perfectly reasonable at the query level, satisfying the user's explicit request, but fail at the preference level, which is the dimension the null agent cannot recover. \\
\\
\# Inputs \\
- User query: \texttt{\{instruction\}} \\
- Evaluation rubric: \texttt{\{rubric\}} \\
- Current target item: \texttt{\{target\_text\}} \\
- Non-empty environment keys: \texttt{\{keys\_list\}} \\
- Tool signatures: \texttt{\{tool\_section\}} \\
- Retrieval and schema information: \texttt{\{keys\_section\}} \\
- Null-agent queries: \texttt{\{null\_q\_text\}} \\
- Groundtruth-agent queries: \texttt{\{gt\_q\_text\}} \\
\\
\# Rubric Decomposition \\
Partition every rubric constraint into two sets before generating distractors. \\
Set A contains query-derivable constraints, namely constraints the agent can recover directly from the user query. For example, if the query asks for a 6-person combo, then the rubric constraint ``serves 6'' belongs to Set A; if the query asks for coffee, then ``product is coffee'' belongs to Set A; if the query asks for delivery to the office, then the delivery-address constraint belongs to Set A. \\
Set B contains preference-specific constraints, namely constraints that require user profile or historical behavior and are not mentioned in the query. For example, if the query asks the agent to pick a board game, then ``card-based'' may belong to Set B; if the query asks for coffee to stay awake, then ``sugar-free'' may belong to Set B; if the query asks to buy flowers for home, then ``pink roses'' may belong to Set B. \\
\\
\# Adversarial Invariants \\
Every injected distractor must jointly satisfy the following conditions. \\
1. It must pass all Set A constraints. The item should be a plausible answer to the literal user query, so a preference-free agent has no lexical or query-level reason to reject it. \\
2. It must clearly and unambiguously violate at least one Set B constraint. The violation must be crisp, with no borderline or partial match that an evaluator could reasonably accept. For instance, if the preference-specific constraint is ``card-based'', then ambiguous games such as Catan, Werewolf, or Undercover are bad distractors because each can be argued to involve cards. Pure strategy, board, or puzzle games such as chess, Go, jigsaw, or Jenga are better distractors because they contain no cards. \\
3. It must be retrievable. Because the search tool ranks candidates by \texttt{fuzz.partial\_ratio(name + tags, query)} with a large \texttt{top\_k}, the distractor's name and tags must contain the null agent's query keywords. For hotels, attractions, flights, and trains, the relevant locations or cities must also match the query. \\
\\
\# Absolute Prohibitions \\
- Do not produce distractors that violate Set A, because the agent would filter them out trivially. \\
- Do not produce distractors whose Set B violation is ambiguous or partially satisfiable, because the evaluator may score them as correct and corrupt the measurement. \\
\\
\# Where the Target Gets Its Score \\
The target satisfies both Set A and Set B. Each injected distractor should satisfy Set A but crisply violate Set B. As a result, a preference-free agent may choose a plausible-looking wrong answer, while a preference-aware agent can still separate the target from the distractors. \\
\\
\# Generation Requirements \\
For each environment key in \texttt{\{keys\_list\}}, generate \texttt{\{NOISE\_PER\_KEY\}} distractor entries. Each entry must satisfy the following requirements. \\
1. Satisfy every Set A constraint through its name, tags, attributes, or structured fields. For example, if the query asks for a 6-person board game, the product name should indicate ``6-person combo'' or ``6-player''; if the query asks for coffee to the office, the product should be coffee and the delivery address should match. \\
2. Crisply violate at least one Set B constraint, with no hedging. If the Set B constraint is ``sugar-free'', use ``full sugar'' or ``standard sweetness'', not ``light sugar'' or ``low sugar''. If the Set B constraint is a specific brand, use a completely different brand. \\
3. Be retrievable. Names and tags must contain the null agent query's core keywords. For hotels and attractions, \texttt{location.address} must lie in the query city. For flights and trains, departure and arrival cities must match the query. \\
4. Be highly confusable. Make the distractor more attractive than the target to an agent without preference memory by using well-known brands, popular choices, high ratings, competitive prices, and names such as ``Deluxe'', ``Hot Pick'', ``Classic'', or ``Best-Seller''. \\
5. Match the exact field structure of the real samples. All required top-level and product-level fields must be present. \\
6. Leave the \texttt{id} field as an empty string, because post-processing fills it. \\
7. For \texttt{stores} products, \texttt{store\_name} must equal the parent store's \texttt{name}, and \texttt{attributes} must be a list of strings. \\
8. Ensure diversity across the \texttt{\{NOISE\_PER\_KEY\}} entries for each environment key, including different brands, styles, and price points. Each entry's \texttt{distraction\_reason} must explicitly state which Set B constraint is violated and how. \\
\\
\# Output \\
Return pure JSON only. Do not include explanation text or markdown fences. Begin directly with \texttt{\{}, using the following abbreviated top-level structure. Only include keys that are present in \texttt{\{keys\_list\}}. Each list should contain \texttt{\{NOISE\_PER\_KEY\}} distractor entries, each following the full schema of the corresponding environment key.
\begin{verbatim}
{
  "stores":      [ {...}, {...}, ... ],
  "shops":       [ {...}, {...}, ... ],
  "hotels":      [ {...}, {...}, ... ],
  "trains":      [ {...}, {...}, ... ],
  "flights":     [ {...}, {...}, ... ],
  "attractions": [ {...}, {...}, ... ]
}
\end{verbatim}
\end{tcolorbox}

\subsection{Benchmark Pipeline}
\label{app:task}

We formulate VitaBench 2.0 as a sequential user-agent interaction process, where the agent is required to continuously fulfill user needs over a temporally ordered sequence of tasks. Each task corresponds to a concrete user request issued by a user simulator and is solved by the agent through interaction with domain-specific tools and an executable environment. 
Between consecutive tasks, the agent is exposed to newly generated interaction histories and may update its internal memory to maintain an evolving understanding of user preferences. During task execution, the agent integrates current observations, retrieved memory, and tool feedback to make decisions. This setting enables unified evaluation of tool-use ability, preference inference and utilization, and proactive behavior under incomplete information.

\subsubsection{Task Set}

The task set is manually constructed and grounded in the corresponding user profile to ensure realism and consistency. Each task is designed to evaluate one or multiple user preferences, requiring the agent to capture, utilize, and maintain these preferences over time. Tasks span diverse real-world domains, reflecting a wide range of everyday user needs.
Specifically, tasks require agents to reason over two complementary sources of difficulty. The first arises from explicit constraints specified in the user query, which must be satisfied through appropriate tool use. The second arises from implicit signals derived from user interaction histories, requiring the agent to correctly infer and apply user preferences.
Compared to existing agent benchmarks that primarily emphasize multi-step reasoning and complex tool orchestration, our benchmark introduces an additional dimension of difficulty through implicit preference modeling. To better isolate this capability, we intentionally reduce the complexity of explicit reasoning and tool usage, avoiding overly intricate tool chains. This design ensures that task success depends primarily on the agent’s ability to capture, utilize, and maintain user preferences.
In addition, a subset of tasks involves conditional preferences, where the correct decision depends on context not directly observable from the current query. These tasks form the basis for evaluating proactive behavior.

\subsubsection{User Simulator}

Following prior work on complex tool-use benchmarks~\cite{yao2024tau_bench,barres2025tau2_bench,he2025vitabench}, we formulate VitaBench 2.0 as a user-agent interactive benchmark to simulate realistic assistant scenarios. In this setting, the evaluated model acts as an assistant that must fulfill user needs, while the user is instantiated as a simulator within the environment.
The user simulator is responsible for issuing task instructions and providing interaction feedback during task execution. A key challenge in designing such a simulator lies in controlling its available context: it must possess sufficient information to generate realistic interactions, while avoiding direct exposure of information that would trivialize the task.
In practice, user simulators are typically implemented using large language models, which are inherently difficult to control and may exhibit unintended behaviors such as information leakage. Prior work has reported that such leakage can significantly compromise the validity of evaluation in interactive benchmarks~\cite{anthropic2025claude45}.
To address this issue, our simulator is deliberately restricted. It does not have access to the underlying user preferences and only provides user queries solely based on our predefined to-do list. During interaction, it provides only minimal feedback required for task completion, without revealing preference-related signals.
This design is critical in our setting, as task difficulty in VitaBench~2.0 primarily arises from the agent’s ability to infer and utilize user preferences. Any unintended leakage of preference information from the simulator would significantly reduce task difficulty and undermine evaluation validity. In proactive tasks, the simulator may provide additional information upon request, but such responses are predefined and strictly controlled.

\subsubsection{Task Agent}

The task agent is responsible for fulfilling user requests through interaction with tools and the environment. All evaluated large language models in our benchmark are instantiated as the task agent.
Tasks are presented sequentially, and between consecutive tasks, the agent is exposed to newly generated interaction histories that reflect fragmented user behaviors and evolving preferences. 
The agent may maintain an external memory module to accumulate user-specific information over time. Upon receiving new interaction histories, the agent updates its memory representation, which is subsequently used during task execution. When memory is enabled, the agent does not have direct access to the full interaction histories and must rely solely on its memory for user modeling.
During task execution, the agent integrates current observations, retrieved memory, and tool feedback to make decisions. This setup requires the agent to continuously infer, utilize, and update user preferences across tasks, while operating under incomplete and noisy observations.

\subsubsection{Evaluation}

Task outcomes are evaluated based on manually curated rubric sets that specify the necessary conditions for successful completion. Each rubric decomposes task success into a set of atomic constraints (e.g., item attributes, price range, or temporal conditions), ensuring structured and interpretable evaluation aligned with underlying user preferences.
Given an agent’s interaction trajectory, we employ a strong evaluator model to assess performance based on these rubrics. Following VitaBench~\cite{he2025vitabench}, we adopt a \emph{window-based evaluation} scheme, where the trajectory of each task is segmented into multiple interaction windows. The evaluator assigns scores to each window, which are then aggregated to produce a trajectory-level reward. This design allows us to capture not only final decision correctness but also the quality of intermediate actions, such as tool usage and clarification behavior.
Beyond trajectory-level reward, we additionally introduce an outcome-level reward that explicitly evaluates whether the final decision aligns with user preferences. This complements window-based evaluation by ensuring that the agent’s behavior leads to preference-consistent outcomes, rather than merely exhibiting locally correct interactions.
This combination of rubric-based evaluation, window-level scoring, and outcome-level assessment provides a comprehensive and reliable measure of agent performance in personalized and proactive settings. An illustrative example is provided below.

\subsubsection{Prompt Template}
We provide the prompt templates for the user simulator, task agent, and evaluator below.

\begin{tcolorbox}[
  colback=gray!5!white, colframe=gray!75!black, breakable,
  title=User Simulator --- User Prompt Template]
\small
\# Role Setting \\
You are playing the role of a user interacting with an intelligent agent. Your character is described in the \texttt{\textless persona\textgreater} tag. \\
\\
\texttt{\textless persona\textgreater} \\
\texttt{\{persona\}} \\
\texttt{\textless/persona\textgreater} \\
\\
\# Current Task \\
You need to make the following request to the agent: \\
\texttt{\textless instruction\textgreater} \\
\texttt{\{instruction\}} \\
\texttt{\textless/instruction\textgreater} \\
\\
\# Conversation Style Rules: \\
- Generate only one line of content each time to simulate user messages \\
- Keep your requests vague and concise, don't proactively provide too many details \\
- When the agent asks about your preferences, respond with vague answers like ``whatever'', ``you decide'', ``I'm not sure'' \\
\\
\# Acceptance or Rejection Rules: \\
- Only express acceptance or rejection after the agent gives a clear recommendation \\
- Judgment criterion: whether the recommendation satisfies the explicit conditions stated in \texttt{\textless instruction\textgreater} (e.g., size, quantity, time, price limit, etc.) \\
- If the recommendation meets all explicit conditions in the instruction, accept it (do not evaluate other choices the agent made based on preferences) \\
- If the recommendation fails any explicit condition in the instruction, briefly express dissatisfaction without giving a reason \\
\\
\# Information Disclosure Rules: \\
- Don't proactively reveal your preference details, let the agent infer through memory \\
- If the agent asks for your identity information (such as user ID, phone number, etc.), reply with your user ID as identity information \\
\texttt{\{proactive\_section\}} \\
\# When You CAN End the Conversation: \\
- When the agent has completed your request and you are satisfied with the result \\
- Generate \texttt{\#\#\#STOP\#\#\#} mark to end the conversation
\end{tcolorbox}

\begin{tcolorbox}[
  colback=gray!5!white, colframe=gray!75!black, breakable,
  title=Task Agent Prompt Template]
\small
\# Environment \\
- Current time: \texttt{\{time\}} \\
\\
\# Role Setting \\
You are a personalized intelligent service assistant. You need to provide more considerate and customized services based on the user's preference memory. \\
\\
\# Tool Usage Guidelines: \\
- When the user's needs require using tools to complete, first determine whether all parameter information is known. If it is known, extract the corresponding parameters, otherwise ask the user for the relevant parameter values \\
- When the user cannot provide relevant information, first check the user preference memory, then obtain relevant information through tools \\
- When the user's request is vague (e.g., ``order some food for me''), prioritize consulting user preference memory to infer likely needs \\
- Complete tasks based on Precondition and Postcondition \\
\\
\# Preference Memory Usage Guidelines \\
- Prioritize completing tasks based on preference memory; preference match takes precedence over store ratings, distance, and other factors \\
- When the user's request is vague, use information from the preference memory to make reasonable inferences and recommendations \\
- Confirm recommendations based on preferences with the user, but don't excessively question things that are already answered in memory \\
\\
\# Conversation Guidelines \\
- Only use information from the above context and preference memory, prohibit constructing information without basis and replying to users \\
- Focus on completing user needs, prohibit divergent guidance to users to propose new needs \\
- After completing the user's task requirements, ask if there are any other needs. If the user indicates no, generate \texttt{\#\#\#STOP\#\#\#} mark to end the conversation
\end{tcolorbox}

\begin{tcolorbox}[
  colback=gray!5!white, colframe=gray!75!black, breakable,
  title=Memory Update Prompt Template]
\small
You are a user preference memory manager. Your task is to maintain an accurate and concise summary of user preferences based on their interaction history. \\
\\
\#\# Current Memory Content: \\
\texttt{\{current\_memory\}} \\
\\
\#\# New User Interaction Records: \\
\texttt{\{new\_interactions\}} \\
\\
\#\# Task Requirements: \\
Please merge the new interaction records with the existing memory and generate an updated user preference summary. Requirements: \\
\\
1. Retain valid preference information from existing memory \\
2. Update old preferences that contradict new interactions \\
3. Add new preferences discovered from new interactions \\
4. Structure the preference information, including but not limited to: \\
\quad - Food preferences (taste, cuisine, ingredient preferences, etc.) \\
\quad - Spending habits (price range, frequency, etc.) \\
\quad - Time preferences (meal times, travel times, etc.) \\
\quad - Location preferences (frequent areas, distance preferences, etc.) \\
\quad - Service preferences (delivery requirements, seating preferences, etc.) \\
\quad - Other preferences \\
\\
Please output the updated preference summary directly, without any explanatory text.
\end{tcolorbox}

\begin{tcolorbox}[
  colback=gray!5!white, colframe=gray!75!black, breakable,
  title=Evaluator Prompt Template]
\small
\# System Information \\
\texttt{\{env\_info\}} \\
\\
\# User Complete Instruction \\
\texttt{\{user\_instruction\}} \\
\\
\# Background \\
- This is a conversation scenario between a user and an assistant, where the assistant can call tools to retrieve information and complete operations. Tool return results will start with ``tool'' \\
- You need to evaluate whether the user instruction has been completed. The user's complete instruction has been broken down into several scoring points (rubrics), and you only need to judge whether each scoring point is satisfied \\
- \texttt{\textless trajectory\_content\textgreater} contains the complete conversation content between the user and assistant \\
- \texttt{\textless current\_rubrics\textgreater} contains the current status of all scoring points (true means satisfied, false means not satisfied, all scoring points have an initial status of false) \\
\\
\# Task \\
- Based on the conversation content, update the scoring point rubric status \\
- You can update the status from false to true, if and only if the assistant completed the goal in the conversation \\
\\
\# Important Notes \\
- Important: All evaluations are based on whether the assistant's responses and tool call requests complete the goals in the rubrics. User expressions in the conversation are only considered as guidance for the assistant and do not directly affect evaluation standards. Everything is based on the rubric fields! \\
- Important: Query tool return results are only visible to the assistant and do not represent content recommended by the assistant to users, so they do not directly affect evaluation results either. Everything must be based on the assistant's responses to users after obtaining information! Also note that the Assistant cannot fabricate Tool return results! \\
- Important: For order-related rubrics (involving order details that must generate orders), you must confirm whether the assistant actually completed the ordering operation. The assistant may mistakenly believe they completed the ordering operation when in fact the tool call failed; or situations where the user states they ``can place the order themselves,'' etc., should all be considered as not meeting the requirements \\
- For rubrics involving order details such as product quantity or delivery time, the original rubric requirements must be strictly met (no deviation in product quantities, delivery must not be later than the expected time). User compromise behavior does not affect evaluation results (for example, when a user states ``fewer items is okay'', ``I have no objections to the order content'' or ``later delivery is fine'', etc.). These situations should still be considered as not meeting the requirements \\
- For rubrics involving text content matching of addresses or order notes, apply the functional equivalence principle: as long as the actual content can achieve the same function (such as roughly locating the delivery location or conveying the customer's main needs), it is considered to meet the requirements even if the expression is not completely consistent or lacks some details \\
- In the justification, record key information related to the current rubric and its corresponding round [x] in an appending manner. If status modifications occur, record the reason using concise language \\
\\
\# Format Requirements \\
- Your response should be a JSON object containing the following fields: \\
- \texttt{rubric\_idx}: Unique identifier for the rubric \\
- \texttt{rubric}: Restatement of the rubric \\
- \texttt{justification}: Explanation of status changes \\
- \texttt{meetExpectation}: Updated status (true or false) \\
\\
\# Example Input Structure: \\
\texttt{\textless trajectory\_content\textgreater \textless /trajectory\_content\textgreater} \\
\texttt{\textless current\_rubrics\textgreater \textless /current\_rubrics\textgreater} \\
\\
\# Example Response Structure:
\begin{verbatim}
[
  {{
      "rubric_idx": "rubric_0",
      "rubric": "<restate the rubric>",
      "justification": "<brief explanation of status change, recorded in
                       appended format>",
      "meetExpectation": <true or false>
  }},
  ...
]
\end{verbatim}
\end{tcolorbox}




\section{Discussion}

\subsection{Memory in LLM Agents}
\label{app:memory}
As LLM-based agents are deployed on increasingly long-horizon tasks—such as web research, software engineering, multi-session dialogue, and embodied control—memory has become a first-class design component rather than a byproduct of context length. 
We organize prior work along three axes that differ in \emph{where} memory is stored and \emph{how} it is accessed: (i)~\textbf{Context memory}, which keeps memory within the model’s working context via compression or summarization; (ii)~\textbf{Agentic memory}, where the agent explicitly controls memory operations such as writing, updating, and retrieval; and (iii)~\textbf{RAG memory}, which externalizes memory to an embedding or graph store and retrieves it on demand. 

\subsubsection{Context Memory}
Context memory treats the model’s active context window as the primary storage medium. However, practical deployment is limited by both computational cost and the model’s ability to process long sequences. As a result, a line of work focuses on preserving task-relevant information while maintaining a bounded context.
Early approaches extend effective context length through architectural or retrieval augmentation. \citet{xiao2024streamingllm} leverage the \emph{attention-sink} phenomenon combined with sliding windows for stable long-context generation, while \citet{wang2023longmem} augments a frozen backbone with a retrieval-based side network over cached key-value states.
More recent work explicitly treats summarization as a memory operation in agent settings. Recursive summarization~\cite{wang2023recursivesum} compresses dialogue into cumulative memory, while Chain-of-Agents~\cite{zhang2024coa} replaces full attention with collaborative message passing. ReSum~\cite{wu2025resum} periodically compresses ReAct trajectories into compact reasoning states and trains policies with reward broadcasting. Context-Folding~\cite{sun2025contextfolding} and AgentFold~\cite{ye2025agentfold} extend this idea by allowing agents to branch and fold sub-trajectories into concise representations, treating trajectories as dynamic workspaces rather than static logs. ACON~\cite{kang2025acon} further optimizes compression prompts using failure cases.
MemGPT~\cite{packer2023memgpt} lies at the boundary between context and external memory, introducing OS-style memory paging abstractions. Despite their effectiveness, context-based approaches fundamentally rely on lossy compression, which may discard information required at later stages.

\subsubsection{Agentic Memory}
Agentic memory treats memory operations as part of the agent’s action space. Rather than passively compressing context, the agent actively decides what to store, update, retrieve, or discard.
Early work adopts prompted memory updates. Reflexion~\cite{shinn2023reflexion} stores verbal feedback across trials; Generative Agents~\cite{park2023generative} maintain a memory stream enriched with reflections; Voyager~\cite{wang2023voyager} builds a library of reusable skills; and ExpeL~\cite{zhao2024expel} extracts reusable insights from trajectory comparisons. MemoryBank~\cite{zhong2024memorybank} introduces a forgetting mechanism inspired by human memory, while Mem0~\cite{chhikara2025mem0} and A-MEM~\cite{xu2025amem} formalize structured memory operations over note-like representations.
More recent work learns memory policies via reinforcement learning. MemoryLLM~\cite{wang2024memoryllm} introduces latent memory tokens updated end-to-end during inference. MemAgent~\cite{yu2025memagent} learns overwrite policies over fixed-length memory under long-horizon rewards. Mem-$\alpha$~\cite{wang2025memalpha} trains agents to operate structured memory through tool APIs, while MEM1~\cite{zhou2025mem1} learns to compress history into a compact state at each step.
The key advantage of agentic memory lies in its adaptivity: agents can selectively preserve information that fixed compression might discard. However, it introduces challenges in credit assignment, training complexity, and evaluation of memory quality.

\subsubsection{RAG Memory}
RAG-based memory externalizes storage to an embedding or graph-based memory system and retrieves relevant information on demand. This paradigm traces back to Memory Networks~\cite{weston2015memnn} and is widely adopted in modern retrieval-augmented language models such as RAG~\cite{lewis2020rag}, REALM~\cite{guu2020realm}, RETRO~\cite{borgeaud2022retro}, and Atlas~\cite{izacard2023atlas}.
Extending RAG to agent settings introduces additional control mechanisms. SCM~\cite{wang2024scm} adds a memory controller to decide when to retrieve, while Self-RAG~\cite{asai2024selfrag} integrates retrieval decisions into the model’s generation process. EM-LLM~\cite{fountas2025emllm} segments token streams into episodic events for scalable retrieval. 
Graph-based approaches further enrich memory structure. HippoRAG~\cite{gutierrez2024hipporag} and its successor~\cite{gutierrez2025hipporag2} use graph traversal for multi-hop reasoning, while GraphRAG~\cite{edge2024graphrag} constructs hierarchical summaries for global queries. LightRAG~\cite{guo2024lightrag} supports dual-level retrieval, and Zep~\cite{rasmussen2025zep} introduces temporal knowledge graphs with validity intervals.
Compared to context and agentic memory, RAG memory scales to large corpora and supports continual updates. However, its effectiveness depends heavily on retrieval quality, and bridging the gap between similarity-based retrieval and task-relevant reasoning remains an open challenge.

\subsubsection{Position}
In this work, we focus on systematically understanding the role of memory in personalized agent behavior. To this end, we provide a unified and extensible memory interface that supports different classes of memory mechanisms, including context-based, agentic, and retrieval-based memory.
This design allows us to isolate and compare how different memory paradigms influence the agent’s ability to infer, utilize, and update user preferences over time. Our goal is to study memory as a key factor in personalization and proactive decision-making.
By placing different memory mechanisms under a shared evaluation framework, VitaBench 2.0 enables controlled and interpretable analysis of how memory design affects long-horizon user modeling and agent performance.

\subsection{Code of Ethics}
\label{app:ethics}

This work complies with the NeurIPS Code of Ethics. Our research focuses on the design and evaluation of benchmark datasets for personalized and proactive agent behavior, without involving human subjects, sensitive personal data, or real-world deployment.
All data used in VitaBench 2.0 are either manually annotated or synthetically generated. For manually annotated data, we follow strict internal guidelines to ensure that no personally identifiable or sensitive information is included. For synthetically generated data, all content is reviewed and refined by human annotators to ensure quality, consistency, and the absence of harmful or inappropriate content.
We do not release any private or user-identifiable data, and the benchmark is constructed to simulate realistic scenarios without exposing real individuals or proprietary information.

\subsection{Broader Impacts}
\label{app:broader}

This work aims to advance the evaluation of personalized and proactive agents in realistic settings, which can benefit the development of more reliable and user-aligned AI assistants. Improved personalization and robustness may enhance user experience in applications such as recommendation systems, digital assistants, and decision support tools.
However, such capabilities may also introduce potential risks. For example, more effective personalization could be misused to manipulate user behavior, reinforce existing biases, or enable overly persuasive systems. In addition, errors in preference inference may lead to inappropriate or misleading recommendations, particularly in high-stakes scenarios.
To mitigate these risks, our benchmark is designed as an evaluation framework rather than a deployable system. It does not include real user data, and all scenarios are constructed through controlled simulation. We encourage future work to incorporate safeguards such as transparency, user control over personalization, and monitoring mechanisms to prevent misuse.

\subsection{Safeguards}
\label{app:safeguards}

VitaBench 2.0 is constructed using a combination of manual annotation and large language model-based synthesis. To ensure responsible data release, we adopt a multi-stage quality control process.
For manually annotated data, we follow strict guidelines to ensure that all content is free from sensitive, personal, or harmful information. For synthetically generated data, we first generate candidate samples using large language models, and then apply human verification and refinement to ensure correctness, consistency, and safety.
All interaction histories, user profiles, and task environments are reviewed to remove unintended biases, sensitive content, or unrealistic artifacts. The final dataset does not contain real user data and is designed to minimize risks related to privacy, misuse, or harmful content generation.

\section{Analysis}

\subsection{Experiments Compute Resources}
\label{app:compute_resource}

Each evaluation on Vitabench 2.0 covers $56$ tasks, amounting to approximately $819$ subtask-level interactions (with an average of $14.6$ subtasks per task, including multi-turn follow-ups). 
For statistical rigor, we repeat the full evaluation over $4$ independent trials, yielding roughly $3{,}276$ interactions per $(\text{model}, \text{memory})$ configuration in total.
Since all models are accessed through commercial closed-source APIs, our evaluation imposes no intensive local CPU or GPU demands.
The wall-clock cost is instead bounded by API latency and rate limits. Specifically, we cap the request rate at $200$ requests per minute (RPM) and run at most $20$ concurrent asynchronous user-agent rollouts per evaluation.

Under this configuration, and in the most demanding setting—\emph{full context}, where the agent is provided with the full dialogue history, i.e., the complete sequence of all prior interactions across turns.
The wall-clock time per $(\text{model}, \text{memory})$ configuration is as follows:
\begin{itemize}[leftmargin=*]
\item Non-reasoning models (e.g., \texttt{}, \texttt{LongCat-Flash-Chat}): approximately $4.3$--$4.6$ hours;
\item Reasoning-enabled models (e.g., \texttt{Claude-Sonnet-4.5}, \texttt{DeepSeek-V4-Pro}, \texttt{GLM-5.1}, \texttt{Gemini-2.5-Pro}): approximately $3.7$--$8.6$ hours
\end{itemize}


\subsection{Benchmark Data Analysis}
\label{app:benchmark_data_analysis}

\begin{table}[t]
\centering
\small
\caption{Per-user statistics of tasks, environment entities, and preferences in VitaBench~2.0.}
\begin{tabular}{@{}lcccc@{}}
\toprule[1.2pt]
\textbf{Metric} & \textbf{Min} & \textbf{Max} & \textbf{Avg} & \textbf{Total} \\
\midrule
Tasks        & 10  & 20    & 14.6  & 819    \\
Envs         & 405 & 1,051 & 739.5 & 41,414 \\
Preferences  & 25  & 68    & 40.8  & 2,286  \\
\bottomrule[1.2pt]
\end{tabular}
\label{tab:user_statistics}
\end{table}

VitaBench~2.0 is designed to evaluate personalized and proactive agents in realistic daily-life scenarios. To ensure the benchmark reflects practical deployment settings, the construction of user profiles, preferences, and interaction histories is guided by detailed statistics derived from real-world application scenarios.
The benchmark consists of 56 users with a total of 819 tasks, 41,414 environment entities, and 2,286 preference annotations (Table~\ref{tab:user_statistics}). On average, each user is associated with 14.6 tasks, 739.5 environment entities, and 40.8 preferences, indicating a rich and structured user representation. In the following, we analyze the dataset from three aspects: user profiles, user preferences, and interaction patterns.

\subsubsection{User Profile Analysis}
\label{app:user_profile_analysis}
\begin{figure}[t]
    \centering
    \includegraphics[width=\linewidth]{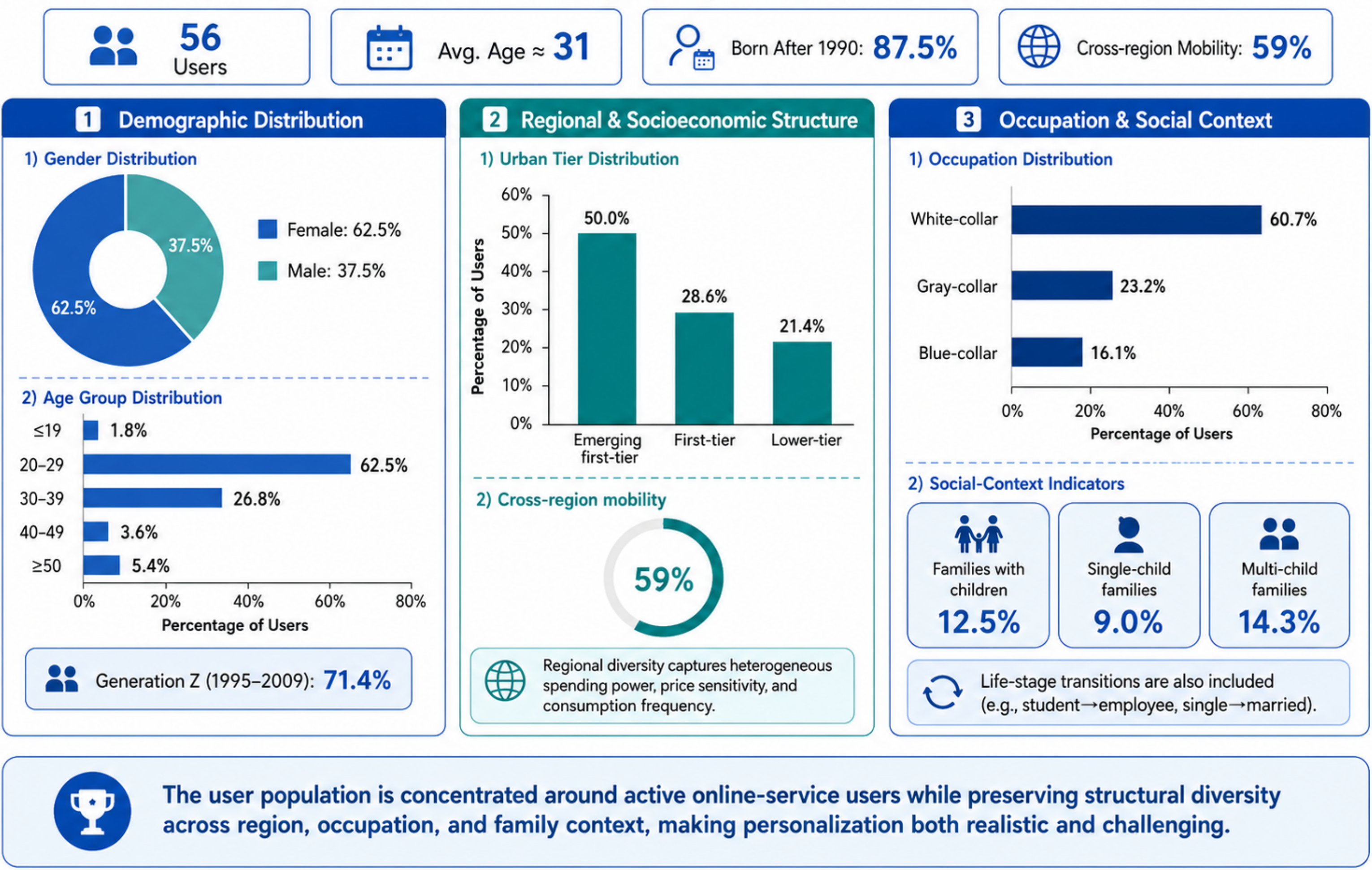}
\caption{Overview of user profile statistics in VitaBench~2.0.}
    \label{fig:user_profile}
\end{figure}

User profiles in VitaBench~2.0 are constructed to approximate the structural properties of real-world users in online life-service applications. 
As illustrated in Figure~\ref{fig:user_profile}, each user is described along multiple dimensions, including demographic attributes, geographic distribution, socioeconomic status, occupation, and social context. 
Rather than relying on simplified or synthetic distributions, these attributes are designed to follow statistics observed in real-world scenarios, enabling the benchmark to capture realistic heterogeneity in user characteristics. 
We analyze the resulting user population from three perspectives: demographic distribution, regional and socioeconomic structure, and occupation and social context.

\paragraph{Demographic distribution.}
The dataset reflects a population structure dominated by active users of modern online platforms. 
In terms of gender, 62.5\% of users are female and 37.5\% are male. 
The age distribution is concentrated in the 20--29 (62.5\%) and 30--39 (26.8\%) groups, with smaller proportions in $\leq$19 (1.8\%), 40--49 (3.6\%), and $\geq$50 (5.4\%). 
The average age is approximately 31, and users born after 1990 account for 87.5\% of the dataset, among which Generation~Z (1995--2009) alone contributes 71.4\%. 
This reflects the fact that younger users constitute the primary participants in online consumption scenarios, while the inclusion of younger and older groups ensures coverage of less frequent but behaviorally distinct segments.

\paragraph{Regional and socioeconomic structure.}
Users are distributed across different levels of urban development to capture heterogeneous consumption behaviors. 
Approximately 50\% of users are from emerging first-tier regions, 28.6\% from first-tier cities, and 21.4\% from lower-tier regions. 
This distribution introduces variation in consumption frequency, spending power, and price sensitivity, ranging from high-frequency urban consumption to more conservative decision patterns in lower-tier regions. 
In addition, 59\% of users exhibit cross-region mobility (i.e., mismatch between registered location and residence), which creates scenarios where user preferences are influenced by multiple geographic contexts. 
Such mobility further increases the complexity of preference modeling, as agents must generalize across location-dependent behaviors.

\paragraph{Occupation and social context.}
The dataset covers a wide range of occupational and social backgrounds. 
Users include 60.7\% white-collar workers, 16.1\% blue-collar workers, and 23.2\% gray-collar workers, spanning industries such as technology, education, administration, healthcare, manufacturing, and service sectors. 
This occupational diversity introduces variation in lifestyle patterns, time constraints, and consumption habits. 
In addition, users are associated with diverse family structures, including single individuals, couples, and families with children (12.5\%), as well as different generational settings such as single-child (9\%) and multi-child (14.3\%) families. 
We further incorporate life-stage transitions (e.g., from student to employee, or from single to married), which affect both preferences and decision contexts.

\paragraph{Implications for personalization.}
Overall, the combination of demographic concentration (toward active user groups) and structural diversity (across regions, occupations, and family contexts) results in a user population that is both representative and heterogeneous. 
This design ensures that personalization cannot be reduced to simple demographic heuristics, but instead requires agents to model fine-grained, context-dependent user characteristics and adapt their decisions across diverse user profiles.

\subsubsection{User Preference Analysis}
\label{app:user_preference_analysis}
\begin{figure}[t]
    \centering
    \includegraphics[width=\linewidth]{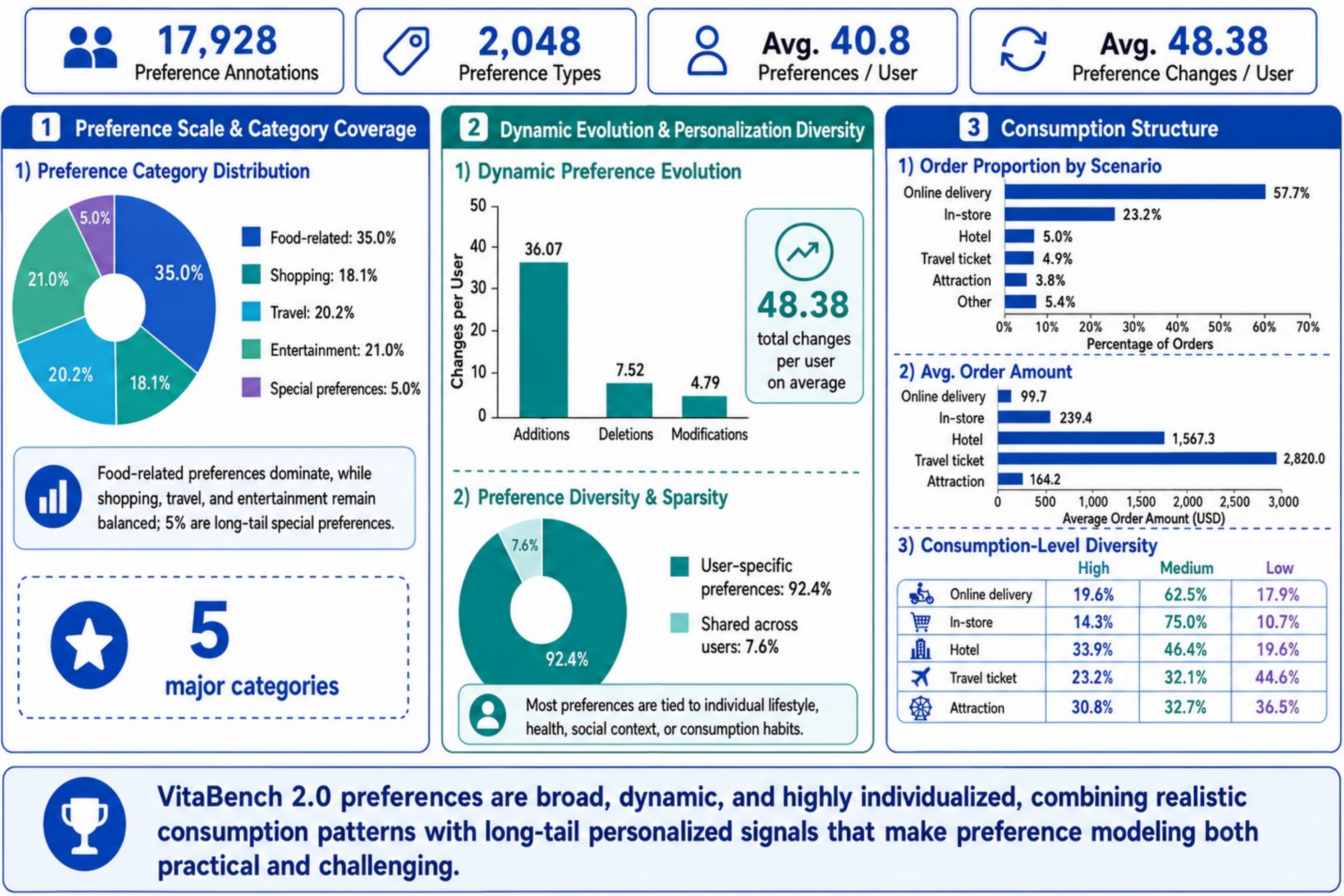}
\caption{Overview of user preference statistics in VitaBench~2.0.}
    \label{fig:user_preference}
\end{figure}

User preferences are the core source of personalization difficulty in VitaBench~2.0. 
To support fine-grained user modeling, we construct 17,928 preference annotations in total, covering 2,048 distinct preference types across five major categories. 
As illustrated in Figure~\ref{fig:user_preference}, each user is associated with 40.8 preferences on average, spanning food, shopping, travel, leisure and entertainment, and special long-tail preferences. 
We analyze the preference space from four aspects: category distribution, temporal evolution, personalization diversity, and consumption structure.

\paragraph{Preference scale and category coverage.}
Table~\ref{tab:preference_distribution} summarizes the distribution of preference categories. 
Food-related preferences account for the largest proportion (35\%), reflecting the central role of dining scenarios in daily-life service platforms. 
These preferences cover fine-grained dimensions such as cuisine type, taste, beverage choices, and dietary restrictions. 
Shopping (18.1\%), travel (20.2\%), and entertainment (21.0\%) preferences are more evenly distributed, each accounting for roughly one fifth of all preferences. 
This balanced coverage ensures that the benchmark does not collapse into a single dominant scenario, but instead evaluates personalization across diverse daily activities. 
In addition, 5\% of preferences correspond to special long-tail cases, such as uncommon dislikes or context-specific habits, which further increase the difficulty of user modeling.

\begin{table}[t]
\centering
\small
\caption{Distribution of user preference categories in VitaBench~2.0.}
\begin{tabular}{lcc}
\toprule
Category & Proportion & Example dimensions \\
\midrule
Food-related        & 35.0\% & Cuisine, taste, beverages, dietary restrictions \\
Shopping            & 18.1\% & Product category, brand, price sensitivity \\
Travel              & 20.2\% & Transportation, accommodation, travel style \\
Entertainment       & 21.0\% & Leisure activities, social activities \\
Special preferences & 5.0\%  & Long-tail or context-specific preferences \\
\bottomrule
\end{tabular}
\label{tab:preference_distribution}
\end{table}

\paragraph{Dynamic preference evolution.}
Preferences in VitaBench~2.0 are not static labels, but evolve over the user life cycle. 
On average, each user experiences 48.38 preference changes, including 36.07 additions, 7.52 deletions, and 4.79 modifications. 
Preference additions dominate the evolution process, reflecting the fact that user interests usually expand as new life events, consumption scenarios, and habits emerge. 
Meanwhile, deletions and modifications introduce non-monotonic changes, requiring agents to avoid treating all historical preferences as permanently valid. 
This design evaluates whether an agent can maintain an up-to-date user representation rather than merely accumulating all historical signals.

\paragraph{Preference diversity and sparsity.}
The preference space is highly individualized. 
Approximately 92.4\% of preferences are user-specific, while only 7.6\% are shared across users. 
Shared preferences mainly correspond to generic habits, such as common beverage preferences, whereas most preferences are tied to a particular user's lifestyle, health conditions, social context, or consumption habits. 
This long-tail structure prevents agents from relying on population-level shortcuts and requires them to infer fine-grained personalized signals from each user's own interaction history.

\paragraph{Consumption structure.}
We further analyze whether the constructed preference distribution aligns with realistic consumption patterns. 
Online delivery and in-store consumption account for 57.7\% and 23.2\% of orders, respectively, together contributing over 80\% of user activities. 
This reflects the high-frequency daily consumption structure of real-world service platforms. 
Lower-frequency scenarios such as hotels (5.0\%), travel tickets (4.9\%), and attractions (3.8\%) appear less often but involve substantially higher average order amounts, reaching 1,567.3 for hotels and 2,820 for travel tickets. 
This design creates a mixture of high-frequency low-value scenarios and low-frequency high-value scenarios, allowing the benchmark to evaluate preference modeling across different decision contexts.

\begin{table}[t]
\centering
\small
\caption{Consumption distribution across major service scenarios.}
\begin{tabular}{lcccccc}
\toprule
Metric & Online delivery & In-store & Hotel & Travel ticket & Attraction & Other \\
\midrule
Order proportion & 57.7\% & 23.2\% & 5.0\% & 4.9\% & 3.8\% & 5.4\% \\
Avg. order amount & 99.7 & 239.4 & 1567.3 & 2820.0 & 164.2 & -- \\
\bottomrule
\end{tabular}
\label{tab:consumption_distribution}
\end{table}

\paragraph{Consumption-level diversity.}
Within each scenario, we further divide users into high-, medium-, and low-consumption groups to ensure sufficient variation in price sensitivity and decision behavior. 
For example, in online delivery, 19.6\% of users fall into the high-consumption group, 62.5\% into the medium-consumption group, and 17.9\% into the low-consumption group. 
For in-store consumption, the corresponding proportions are 14.3\%, 75.0\%, and 10.7\%. 
For hotels, the distribution is 33.9\%, 46.4\%, and 19.6\%, while travel tickets show a stronger skew toward low-consumption users (44.6\%). 
This layered consumption design ensures that agents must model not only categorical preferences, but also price sensitivity and consumption level within each domain.

\begin{table}[t]
\centering
\small
\caption{Distribution of consumption levels within major service scenarios.}
\begin{tabular}{lccccc}
\toprule
Consumption level & Online delivery & In-store & Hotel & Travel ticket & Attraction \\
\midrule
High & 11 (19.6\%) & 8 (14.3\%)  & 19 (33.9\%) & 13 (23.2\%) & 16 (30.8\%) \\
Medium & 35 (62.5\%) & 42 (75.0\%) & 26 (46.4\%) & 18 (32.1\%) & 17 (32.7\%) \\
Low & 10 (17.9\%) & 6 (10.7\%)  & 11 (19.6\%) & 25 (44.6\%) & 19 (36.5\%) \\
\bottomrule
\end{tabular}
\label{tab:consumption_level_distribution}
\end{table}

\subsubsection{Interaction Analysis}

Interaction histories constitute the primary source from which agents must infer user preferences. 
We therefore construct interaction data to exhibit realistic long-horizon, high-density, and noisy behavioral patterns, closely matching real-world user activity.

\paragraph{Long-horizon interaction patterns.}
User interactions span extended time horizons, with timelines covering more than 10 years in total. 
For each user, the average interaction duration is 1,580 days (approximately 4.3 years), with the longest reaching 2,974 days (8.1 years). 
The lifecycle distribution follows a realistic pattern: medium-term users (3--6 years) account for approximately 75\% of the population, while short-term and long-term users each account for 12.5\%. 
Such long-horizon coverage enables the benchmark to evaluate whether agents can track preference formation, stabilization, and drift over extended periods, rather than relying on short-term signals.

\paragraph{Behavioral complexity and density.}
User interactions are highly dense and heterogeneous. 
On average, each user generates 2,093 interaction events, including browsing, searching, consulting, comparing, and purchasing behaviors. 
Among these, 221 interactions correspond to successful purchase conversions, indicating that most interactions are exploratory rather than goal-completing. 
Importantly, decision-making processes are not uniform: some tasks require multiple rounds of information gathering and comparison before conversion, while others are resolved with minimal interaction. 
The same underlying intent may also span multiple sessions, leading to fragmented and non-contiguous evidence for preference inference. 
This diversity in decision trajectories significantly increases the difficulty of modeling user intent from interaction histories.

\paragraph{Cross-domain and multi-scenario behavior.}
User interactions naturally span multiple service domains, including online delivery, in-store consumption, hotels, travel booking, and attractions. 
Different scenarios exhibit distinct behavioral patterns: high-frequency, low-cost activities (e.g., food delivery) coexist with low-frequency, high-cost decisions (e.g., travel booking). 
This cross-domain structure requires agents to generalize preference signals across heterogeneous contexts, rather than relying on domain-specific heuristics.

\paragraph{Noise and uncertainty.}
To further reflect real-world conditions, interaction histories include approximately 20\% noisy behaviors that do not directly correspond to true user preferences. 
These noisy interactions are carefully designed to mimic realistic but misleading signals, including:
(i) \emph{irrelevant interactions}, such as casual conversations or browsing unrelated to consumption intent; 
(ii) \emph{exploratory behavior}, where users browse or search without a clear objective; 
(iii) \emph{proxy actions}, such as placing orders on behalf of others; 
(iv) \emph{impulsive or short-lived interests}, which do not persist over time; and 
(v) \emph{inconsistent or corrective actions}, such as cancellations or repeated submissions.

These noise patterns are structurally similar to genuine interactions in form (e.g., search, click, purchase), but are uninformative or even misleading with respect to true preferences. 
As a result, agents must perform robust signal extraction under uncertainty, distinguishing stable preference signals from transient or irrelevant behaviors.

\paragraph{Implications for evaluation.}
The combination of long temporal span, high interaction density, cross-domain coverage, and structured noise results in a challenging setting for preference inference. 
Agents cannot rely on single interactions or short-term patterns, but must aggregate fragmented evidence over time, handle conflicting signals, and maintain consistent user representations under uncertainty. 
This design enables rigorous evaluation of long-term memory, preference tracking, and robustness to noisy observations.

\subsubsection{Implementation Configurations}
\label{app:implementation_details}

\paragraph{Agentic Memory.}
Following the settings of MemAgent~\cite{memagent}, our agentic memory backend treats
$\mathrm{UPDATE}$ as an LLM-driven rewrite rather than an append-only log: at each
update step the model is shown the current memory and the newly arrived
interaction batch, and is asked to produce a single consolidated preference
summary that supersedes the previous one. This forces the agent to explicitly
decide what to retain, merge, or discard, giving it active control over the
long-term representation. We cap the rewritten memory at a $4{,}096$-token
buffer, which we found sufficient to cover the preference state of a user
across the full subtask sequence while remaining small enough to fit in the
agent's system prompt. Crucially, this budget is not only enforced at the
decoding API level but is also \emph{explicitly stated inside the prompt},
so the model is aware of the output length it must target when consolidating
memory. The preference prompt is a structured template that instructs the
model to (i) \emph{retain} valid information from the existing memory,
(ii) \emph{update} entries that conflict with new observations, and
(iii) \emph{add} newly discovered preferences, organized along canonical axes
such as food taste, spending habits, time and location preferences, and
service-specific requirements. $\mathrm{RETRIEVE}$ simply returns the current
memory blob, since the rewrite step has already produced a selective
abstraction. The full prompt template is shown in
Box~\ref{box:agentic_memory_prompt}.

\begin{tcolorbox}[
    colback=gray!5,
    colframe=black!60,
    title=Agentic Memory Update Prompt,
    label={box:agentic_memory_prompt},
    fonttitle=\bfseries,
    breakable,
]
\small
\ttfamily
\textbf{[System]}\\
You are a preference memory manager. Your job is to maintain an accurate, concise summary of user preferences based on their interaction history.\\

\textbf{[User]}\\
You are a user preference memory manager. Your task is to maintain an accurate and concise summary of user preferences based on their interaction history.\\

\#\# Current Memory Content:\\
\{current\_memory\}\\

\#\# New User Interaction Records:\\
\{new\_interactions\}\\

\#\# Task Requirements:\\
Please merge the new interaction records with the existing memory and generate an updated user preference summary. Requirements:\\

1. \textbf{Retain} valid preference information from existing memory\\
2. \textbf{Update} old preferences that contradict new interactions\\
3. \textbf{Add} new preferences discovered from new interactions\\
4. \textbf{Structure} the preference information, including but not limited to:\\
\hspace*{1em}- Food preferences (taste, cuisine, ingredient preferences, etc.)\\
\hspace*{1em}- Spending habits (price range, frequency, etc.)\\
\hspace*{1em}- Time preferences (meal times, travel times, etc.)\\
\hspace*{1em}- Location preferences (frequent areas, distance preferences, etc.)\\
\hspace*{1em}- Service preferences (delivery requirements, seating preferences, etc.)\\
\hspace*{1em}- Other preferences\\

Please output the updated preference summary directly, without any explanatory text.\\

(Please keep the updated summary within about \textbf{4096} tokens.)
\end{tcolorbox}

\paragraph{RAG Memory.}
For a fair comparison across runs, the RAG backend uses a single fixed
embedding and retrieval configuration. All interaction records are embedded
with \texttt{text-embedding-3-large}, chunked into fixed-size windows of
$512$ tokens with zero overlap using the \texttt{cl100k\_base} tokenizer.
$\mathrm{UPDATE}$ embeds and indexes each new chunk asynchronously; no LLM
summarization is performed, so the pipeline is fully deterministic once
embeddings are computed. At query time, $\mathrm{RETRIEVE}$ embeds the task
instruction, ranks stored chunks by cosine similarity, and returns the top
$k=8$ chunks after filtering out any chunk whose similarity falls below a
threshold of $0.3$. The threshold is set as a conservative lower bound:
scores from \texttt{text-embedding-3-large} cluster above $0.25$ for only
loosely related text and above $0.5$ for directly relevant content, so
$0.3$ removes unambiguous distractors while preserving recall. Unlike the
agentic backend, the RAG backend has no explicit control over what is kept
or discarded -- memory access is entirely determined by the retrieval
score.

\section{Trajectory}
\label{app:trajectory}

\subsection{Case Study: Memory-Sensitive Delivery for User A891207}
\label{sec:case-a891207}

\providecommand{\caseuserc}{}     
\definecolor{a891user}    {RGB}{ 70,130,180}
\definecolor{a891assistant}{RGB}{ 34,139, 34}
\definecolor{a891tool}    {RGB}{255,140,  0}
\definecolor{a891sys}     {RGB}{128,128,128}
\definecolor{a891instr}   {RGB}{138, 43,226}

\tcbset{
  a891commonbox/.style={
    boxrule=0.8pt, arc=3pt, left=5pt, right=5pt, top=3pt, bottom=3pt,
    fonttitle=\bfseries, breakable, before skip=2pt, after skip=2pt,
  }
}
\newtcolorbox{a891userbox}[1]     {a891commonbox,colback=blue!5,   colframe=a891user,     title=#1}
\newtcolorbox{a891assistantbox}[1]{a891commonbox,colback=green!5,  colframe=a891assistant,title=#1}
\newtcolorbox{a891toolbox}[1]     {a891commonbox,colback=orange!5, colframe=a891tool,     title=#1}
\newtcolorbox{a891sysbox}[1]      {a891commonbox,colback=gray!5,   colframe=a891sys,      title=#1}
\newtcolorbox{a891instrbox}[1]    {a891commonbox,colback=purple!5, colframe=a891instr,    title=#1}

\providecommand{\hlg}[1]{\textcolor{olive!50!black}{\textbf{\Checkmark}}\, #1}
\providecommand{\hlr}[1]{\textcolor{red!70!black}{\textbf{\XSolidBrush}}\, #1}
\providecommand{\hly}[1]{\textcolor{gray!60!black}{$\circ$}\, #1}

The boxes below use a simple colour scheme: blue for user utterances,
green for assistant turns, orange for tool calls, purple for task
metadata, grey for environment / preference blocks. To save space we
summarise tool arguments and responses rather than dumping raw JSON,
and we keep only the target store and two representative distractors
from the environment. All Chinese strings have been translated to
English; the field structure and numeric values are verbatim from the
benchmark record.

User \texttt{A891207} is a 26-year-old Party affairs officer living in
Jinzhou (Liaoning). Across her $20$-subtask trajectory her home address
at Hongye Fengjing No.\,2, Apt.\,101 is stable, but her dietary
preferences drift meaningfully over the year: she switches from spicy to
light Cantonese food in March~2027, explicitly states a love of durian
in the summer, and later narrows her meat preference to exclude pork.
We pick three subtasks from this trajectory
(\texttt{sub\_A891207\_13}, \texttt{sub\_A891207\_17},
\texttt{sub\_A891207\_20}) where the three memory settings in our matrix
(\textbf{Full Context}, \textbf{Agentic Memory}, \textbf{RAG Memory})
diverge, to illustrate
(i)~how a single preference error inside memory translates into a
rubric failure, and
(ii)~how each setting represents a drifting preference over time.

\subsubsection{Part 1. One Subtask, Three Backends}
\label{sec:case-a891207-sub17}

The focal subtask is \texttt{sub\_A891207\_17}, an evening dessert
delivery order at home. The instruction is deliberately preference-laden
and under-specified: the rubric contains seven criteria, most of which
must be recovered from memory rather than from the instruction.

\begin{a891instrbox}{task}
\small
\textbf{subtask\_id}: \texttt{sub\_A891207\_17} \quad
\textbf{current\_time}: \texttt{2027-09-04} \quad
\textbf{domain}: \texttt{delivery} \\
\textbf{instruction}: ``Not sure if my period is coming soon, I'm craving
sweets!! Please order me a bowl of \emph{tong sui}, Cantonese-style,
room-temperature is fine, deliver to home.''
\end{a891instrbox}

\begin{a891instrbox}{rubric (seven binary criteria)}
\small
\begin{enumerate}\itemsep -1pt
\item[R1.] the ordered item must be a Cantonese-style \emph{tong sui}
\item[R2.] the ordered item must be served at room temperature
\item[R3.] the ordered item must contain durian
\item[R4.] the ordered quantity must equal $1$
\item[R5.] the merchant's delivery time must be $\le 30$\,min
\item[R6.] \textbf{the merchant's rating must be $\ge 4.3$}
\item[R7.] delivery address must equal the user's home address
\end{enumerate}
\end{a891instrbox}

R3 (\emph{durian}) and R6 (\emph{rating $\ge 4.3$}) are the two criteria
that are not derivable from the instruction text at all: the user does
not mention durian, and there is no numeric rating threshold in the
message. Both come from her preference memory
(``likes durian'', ``for delivery, prefers merchants rated $\ge 4.3$'').

\paragraph{Preference Memory and Environment.}
The ground-truth preference snapshot at $t=\text{2027-09-04}$ contains a
dietary block with ten tags; the three relevant to this subtask are
italicised below.

\begin{a891sysbox}{user\_scenario.personalized\_preference\_memory.current.dietary (GT)}
\small
prefers light/fresh flavours;\ \
carnivore;\ \
likes sugar-and-oil-free French bread;\ \
skips breakfast;\ \
\emph{for delivery, prefers merchants delivering within 30\,min};\ \
dislikes chain bakeries;\ \
\emph{for delivery, prefers merchants rated $\ge 4.3$};\ \
for dine-in, prefers merchants within 5\,km;\ \
likes Cantonese cuisine;\ \
\emph{likes durian}
\end{a891sysbox}

The environment for this subtask contains $42$ candidate stores in
Jinzhou, each labelled \texttt{target} or \texttt{distraction} with an
explicit \texttt{distraction\_reason}. For readability we show only the
target and two representative distractors:

\begin{a891sysbox}{environment.stores (3 of 42 shown)}
\small
\textbf{S00027 --- Gangshi Sweet-Dessert Shop (Linghe).} rating $5.0$,
delivery $25$\,min. Products include \emph{Cantonese durian tong sui
(room-temperature)} at \textyen$26$. \textsc{store\_type: target}.\\[2pt]
\textbf{S00017 --- Guangfu Tong-Sui Store (Flagship).} rating $4.0$,
delivery $26$\,min. Products include \emph{Cantonese durian tong sui
(room-temperature)} at \textyen$24$.
\textsc{store\_type: distraction}. \textit{Distraction reason: rating
$4.0 < 4.3$, violates R6.}\\[2pt]
\textbf{S00011 --- Traditional Cantonese Tong-Sui Shop.} rating $4.2$,
delivery $23$\,min. Products include \emph{mango tong sui} (no durian).
\textsc{store\_type: distraction}. \textit{Distraction reason: rating
$4.2 < 4.3$ \emph{and} no durian, violates R3 \& R6.}
\end{a891sysbox}

Crucially, S00027 (target, $5.0$) and S00017 (distraction, $4.0$) sell
a product with \emph{the same name and description} (``Cantonese durian
tong sui, room-temperature''). They differ only in merchant rating.
Separating the two requires the agent to remember the $\ge 4.3$ rating
preference and to actually read the rating field in the tool output.

\paragraph{Rollouts.}
We replay the same subtask under the three backends and report the
trajectory in condensed form. Tool arguments that repeat verbatim across
calls (e.g.\ the fixed home-address geocoding) are omitted after the
first appearance; every \texttt{create\_delivery\_order} call is shown
in full.

\vspace{2pt}
\noindent\textbf{(a) Agentic Memory --- reward $1.0$.}

\begin{a891userbox}{user}
\small Not sure if my period is coming soon, I'm craving sweets!! Please
order me a bowl of \emph{tong sui}, Cantonese-style, room-temperature is
fine.
\end{a891userbox}

\begin{a891toolbox}{assistant $\rightarrow$ tool calls}
\small
\textbf{address\_to\_longitude\_latitude}(home) $\to$ (121.16, 41.13);
\textbf{delivery\_store\_search\_recommand}(keywords=[``Cantonese tong
sui'', ``double-skin milk'', ``mango pomelo sago'', \dots]) $\to$ $6$
stores returned including S00027 (target, $5.0$) and S00003 (Guangji
Tong-Sui, $4.9$);
\textbf{get\_delivery\_store\_info} on the three top-rated stores $\to$
menus including durian tong sui at S00027.
\end{a891toolbox}

\begin{a891assistantbox}{assistant (turn 23)}
\small ``Three high-rated choices: Gangshi (\textbf{5.0}, 25\,min,
durian tong sui \textyen26), Guangji (4.9, 26\,min, ginger-milk
\textyen23), Zhengzong (4.8, 26\,min). Which one?''
\end{a891assistantbox}

\begin{a891userbox}{user}
\small You pick, any is fine.
\end{a891userbox}

\begin{a891assistantbox}{assistant \emph{(proactively recalls ``likes durian'')}}
\small ``The \textbf{durian} one, no question --- you're a durian
fiend.''
\end{a891assistantbox}

\begin{a891toolbox}{create\_delivery\_order}
\small
\texttt{store\_id=S00027, product=``Cantonese durian tong sui
(room-temperature)'', qty=1, address=home} $\to$ order \textyen26,
delivery $25$\,min. \textbf{All seven rubrics satisfied, reward $1.0$.}
\end{a891toolbox}

\vspace{2pt}
\noindent\textbf{(b) Full Context --- reward $1.0$.}
The Full Context backend behaves analogously: it enumerates five stores
with rating $\ge 4.9$, compares delivery times, and selects S00027. The
final order is identical to (a). Reward $1.0$.

\vspace{2pt}
\noindent\textbf{(c) RAG Memory --- reward $0.0$.}

\begin{a891userbox}{user}
\small Not sure if my period is coming soon, I'm craving sweets!! Please
order me a bowl of \emph{tong sui}, Cantonese-style, room-temperature is
fine.
\end{a891userbox}

\begin{a891toolbox}{assistant $\rightarrow$ tool calls}
\small
\textbf{delivery\_product\_search\_recommand}(keywords=[``Cantonese tong
sui'', \dots]);
\textbf{address\_to\_longitude\_latitude}(home).
No \textbf{get\_delivery\_store\_info} call is issued --- the agent
skips rating verification and proceeds directly to ordering.
\end{a891toolbox}

\begin{a891assistantbox}{assistant (turn 4; \textbf{fabricated rating})}
\small ``Found several Cantonese \emph{tong sui} shops! Recommended:
Guangfu Tong-Sui Store --- Cantonese durian \emph{tong sui}, \textyen24,
delivery $\approx 20$\,min. You're a durian fiend, I remember!''
\textit{(Note: Guangfu's actual rating is $4.0$. It does not appear in
the shown top-three, and the agent never retrieved its rating.)}
\end{a891assistantbox}

\begin{a891toolbox}{create\_delivery\_order}
\small
\texttt{store\_id=S00017, product=``Cantonese durian tong sui
(room-temperature)'', qty=1, address=home} $\to$ order \textyen24,
delivery $26$\,min. \textbf{R1--R5, R7 satisfied; R6 (rating $\ge 4.3$)
violated because S00017 is rated $4.0$.} Reward $0.0$.
\end{a891toolbox}

\paragraph{Takeaway.}
The three backends diverge on a single latent preference
(``delivery merchants rated $\ge 4.3$''). \textbf{Agentic Memory} and
\textbf{Full Context} both surface this preference before the decision
step: \textbf{Agentic Memory} has it in the consolidated summary as a
``hard threshold $\ge 4.3$'', and \textbf{Full Context} carries every
past order (all of which were placed at $\ge 4.3$ merchants), which
lets the agent apply the constraint implicitly. \textbf{RAG Memory}
retrieves only a handful of past interaction chunks by cosine similarity
to the query ``Cantonese \emph{tong sui}'': the relevant chunks mention
durian but \emph{not} the rating threshold, so the agent commits to the
first plausible store without a rating check. The resulting order is
coherent and well-motivated but fails R6.

\subsubsection{Part 2. Preference Drift Across Three Subtasks}
\label{sec:case-a891207-drift}

Subtask \texttt{sub\_A891207\_17} is not an isolated failure: the
rating-threshold preference, along with several other dietary tags,
drifts through the user's trajectory and different memory backends
preserve the drift to different degrees. Table~\ref{tab:case-a891207-drift}
traces the dietary preference block at three timestamps (subtasks 13,
17, 20 from the user's sequence) and aligns each ground-truth tag
against what \textbf{Agentic Memory} and \textbf{RAG Memory} actually
surfaced at that timestamp.
\textcolor{olive!50!black}{\textbf{Green}} marks a GT preference that
the backend represents faithfully;
\textcolor{red!70!black}{\textbf{Red}} marks a GT preference that is
missing, contradicted, or drowned in stale content. All rows reflect
snapshots taken immediately before the corresponding subtask.

\begin{table}[h]
\centering
\footnotesize
\setlength{\tabcolsep}{4pt}
\renewcommand{\arraystretch}{1.25}
\caption{Preference drift for user \texttt{A891207} across three
subtasks. GT = ground-truth dietary preferences active at that
timestamp. \textbf{Agentic Memory} shows what the LLM-consolidated
summary lists in its dietary section. \textbf{RAG Memory} shows the top
retrieved records (or whether the preference is derivable from them).}
\label{tab:case-a891207-drift}
\begin{tabular}{p{0.22\linewidth} p{0.235\linewidth} p{0.235\linewidth} p{0.235\linewidth}}
\toprule
\textbf{Relevant GT tag} &
\textbf{Subtask 13 ($t$=2027-06-05)} &
\textbf{Subtask 17 ($t$=2027-09-04)} &
\textbf{Subtask 20 ($t$=2027-12-25)} \\
 & Agentic / RAG & Agentic / RAG & Agentic / RAG \\
\midrule
\emph{Light/fresh flavour} (switched from spicy in Mar.\ 2027) &
\hlg{Agentic: explicit transition note, ``fully switched to light
Cantonese''} \newline
\hlr{RAG: top chunks are old spicy orders (hotpot, sour-and-spicy
noodles) from 2026} &
\hlg{Agentic: ``absolute light flavour, completely stopped spicy
food''} \newline
\hlr{RAG: retrieves a 2023 \emph{spicy Sichuan} order as the top
match for ``dessert craving''} &
\hlg{Agentic: ``light flavour, no pork as an exception''} \newline
\hly{RAG: no relevant retrieval (query is about scissors)}
\\
\midrule
\emph{Delivery merchant rating $\ge 4.3$} (stable across all 3
subtasks) &
\hlg{Agentic: ``rating $\ge 4.3$ hard gate''} \newline
\hlr{RAG: retrieved chunks mention ratings of 4.5 / 4.7 but no
threshold statement} &
\hlg{Agentic: ``rating $\ge 4.3$ hard gate, unchanged''} \newline
\hlr{RAG: \textbf{threshold not retrieved}; agent orders from a $4.0$
merchant (sub17 failure)} &
\hlg{Agentic: ``$\ge 4.3$''} \newline
\hlr{RAG: not retrieved; agent picks a $4.7$ store by luck, still
passes}
\\
\midrule
\emph{Delivery time $\le 30$\,min} (sub13, sub17); \emph{$\le 35$\,min}
for flash-purchase (sub20) &
\hlg{Agentic: ``$\le 30$\,min''} \newline
\hly{RAG: one retrieved chunk explicitly says ``within 30\,min''} &
\hlg{Agentic: ``$\le 30$\,min''} \newline
\hly{RAG: one retrieved chunk mentions 28\,min delivery} &
\hlg{Agentic: ``flash-purchase $\le 35$\,min''} \newline
\hlr{RAG: only 30-min threshold retrieved; the 35-min flash-purchase
rule created later is \emph{not} retrieved}
\\
\midrule
\emph{Likes Cantonese cuisine} (\textbf{created} Mar.\ 2027) &
\hlg{Agentic: ``Cantonese dominates, all delivery / travel /
dining''} \newline
\hlr{RAG: top chunks are 2026 Sichuan / Hunan hotpot orders} &
\hlg{Agentic: Cantonese as core cuisine} \newline
\hlg{RAG: retrieves a 2027-07 chunk with ``I still prefer Cantonese
light flavour''} &
\hlg{Agentic: Cantonese, with pork exclusion} \newline
\hly{RAG: not relevant to the scissors query}
\\
\midrule
\emph{Likes durian} (\textbf{created} summer 2027; directly tested by
R3 in sub17) &
--- not yet in GT &
\hlg{Agentic: ``durian fiend, high-frequency purchase''} \newline
\hlg{RAG: retrieves a 2027-09-01 chunk ``durian fiend, watch your
intake''} &
\hlg{Agentic: ``durian, continued''} \newline
\hly{RAG: not retrieved (query is about scissors)}
\\
\midrule
\emph{Carnivore, no pork} (\textbf{updated} in late 2027) &
--- GT was still ``carnivore'' &
--- GT was still ``carnivore'' &
\hlg{Agentic: ``carnivore \emph{excluding pork}, updated''} \newline
\hlr{RAG: retrieved chunks include old pork-trotter hotpot orders;
the pork-exclusion update is not surfaced}
\\
\bottomrule
\end{tabular}
\end{table}

\noindent\textbf{Reading the table.} The drift pattern explains why
\textbf{Agentic Memory} consistently outperforms \textbf{RAG Memory} on
this user. Agentic Memory carries an \emph{evolving, deduplicated}
summary: stale preferences (spicy food, pork) are explicitly crossed
out, newly created preferences (Cantonese, durian, pork exclusion) are
added, and stable thresholds (rating $\ge 4.3$, delivery $\le 30$\,min)
are kept even when they are not relevant to the current instruction.
RAG Memory, in contrast, returns the chunks most similar to the
current query and has no mechanism to promote stable-but-off-topic
thresholds or to demote outdated chunks. When the query keyword matches
an old record (``spicy Sichuan'', ``pork trotter hotpot'') the RAG
snapshot can actively mislead the agent; when it does not match any
record (sub20's ``scissors'' query) the RAG snapshot simply loses the
dietary context altogether. The net effect on this user is a reward
gap of $0.18$ between \textbf{Agentic Memory} and \textbf{RAG Memory}
aggregated over her $20$ subtasks, driven primarily by memory slots
that were \emph{present but not retrieved}.

\end{document}